\documentclass{article}

\usepackage[preprint]{neurips_2026}

\usepackage[utf8]{inputenc} % allow utf-8 input
\usepackage[T1]{fontenc}    % use 8-bit T1 fonts
\usepackage{hyperref}       % hyperlinks
\usepackage{url}            % simple URL typesetting
\usepackage{booktabs}       % professional-quality tables
\usepackage{amsfonts}       % blackboard math symbols
\usepackage{nicefrac}       % compact symbols for 1/2, etc.
\usepackage{microtype}      % microtypography
\usepackage{xcolor}         % colors
\usepackage{amsmath}
\usepackage{graphicx} 
\usepackage{multirow}
\usepackage{amssymb}
\usepackage{mathtools}
\usepackage{amsthm}
\usepackage{xspace}
\usepackage{setspace}
\usepackage{url}
\usepackage{nicefrac}
\usepackage{booktabs}
\usepackage{amsfonts}
\usepackage{bold-extra}
\usepackage[table]{xcolor}
\usepackage{makecell}
\usepackage{array}

\usepackage{tabularx}
\usepackage{dcolumn,caption}
\usepackage{arydshln}
\usepackage{tikz}
\usepackage{pgfplots}
\usepackage{framed}
\usepackage{pifont}
\usepackage{bbm}
\usepackage{enumitem}
\usepackage{subcaption}
\usepackage{amssymb}
\usepackage{adjustbox}
\usepackage{mdframed}
\usepackage{wrapfig}
\usepackage{float}
\usepackage[most,breakable]{tcolorbox}
\usepackage{verbatim}
\usepackage{booktabs}
\usepackage{multirow}
\usepackage[table]{xcolor}
\usepackage{pifont}
\usepackage{wrapfig}
\usepackage[dvipsnames]{xcolor}
\definecolor{green}{RGB}{36, 214, 36}
\definecolor{red}{RGB}{235, 30, 30}
\usepackage{fontawesome}
\usepackage{listings}
\usepackage{xcolor}

\title{VeriTrip: A Verifiable Benchmark for Travel Planning Agents over Unstructured Web Corpora}

\author{
  \textbf{Yuting Xu}$^{1,2,3}$\thanks{Equal contribution.}, \textbf{Jiayi Tian}$^{3}$\footnotemark[1], \textbf{Jian Liang}$^{4,5}$, \textbf{Xin Xiong}$^{1,2}$, \\
  \textbf{Hang Zhang}$^{3}$, 
  \textbf{Mu Xu}$^{3}$ ,\textbf{Xiao-Yu Zhang}$^{1,2}$\thanks{Corresponding author.}\\
  $^{1}$Institute of Information Engineering, CAS \quad
  $^{2}$School of Cyber Security, UCAS  \\
  $^{3}$Amap, Alibaba Group \quad $^{4}$NLPR \& MAIS, Institute of Automation, CAS  \\
  $^{5}$School of Artificial Intelligence, UCAS\\
  \faGithub\; 
  \href{https://github.com/rainy-xu/VeriTrip}{VeriTrip}
}

\begin{document}

\maketitle

\begin{abstract}
Existing benchmarks have laid the foundation for travel planning agents by establishing API-centric paradigms. 
However, as the capabilities of Autonomous Agents continue to advance, their evaluation must evolve beyond simple tool execution toward handling the inherent complexities of the open web. 
Current benchmarks bypass core cognitive hurdles: they fail to account for information noise, ignore multi-source factual contradictions, and overlook the necessity of grounding visual perception into logical planning. 
We introduce VeriTrip, a verifiable benchmark designed to meet the increasing demands for agent robustness and reliability. VeriTrip shifts the evaluation focus to evidence-grounded reasoning over unstructured multimodal web corpora. 
It establishes a Multimodal Retrieval Base (MRB) derived from real-world sources, forcing agents to autonomously orchestrate queries across heterogeneous data. 
A synchronized Verifiable Knowledge Base (VKB) enables a cell-wise verification protocol that precisely quantifies factual reliability, distinguishing systematic reasoning failures from parametric hallucinations. 
Our evaluations across leading MLLMs reveal a critical \textit{retrieval-reasoning trade-off}: the cognitive load of autonomous retrieval significantly erodes instruction retention. 
VeriTrip provides the rigorous foundation necessary for the next generation of planning agents capable of operating in unconstrained, multimodal environments.

\end{abstract}
%要在方法中突出模型会根据目前搜索到的信息挑战搜索query
\section{Introduction}
\label{sec:intro}
The evolution of autonomous agents is shifting toward complex, long-horizon planning tasks~\cite{li2025review}.
A truly intelligent agent is expected to independently decompose complex objectives, navigate heterogeneous information environments, and synthesize multi-constrained decisions from fragmented evidence. 
This paradigm, often characterized as Retrieval-Augmented Planning~\cite{zare2024rap}, requires an agent to bridge the gap between open-ended uncertainty and verifiable action. 
As agents are increasingly deployed in unconstrained web environments, their success depends not only on the depth of their information gathering but also on the factual integrity and logical consistency of their final outputs.

A fundamental gap persists between existing research-oriented and planning-oriented benchmarks. 
Current deep research benchmarks~\cite{du2025deepresearch,narayan2025deepmmsearch} primarily focus on the breadth of information seeking but lack a rigorous protocol for verifying the resulting synthesis. 
Their evaluations frequently rely on LLM-as-Judge grading for unstructured reports, which is often insufficient to detect unfounded claims or subtle errors introduced during multi-source aggregation~\cite{zhang2025deep}. 
Consequently, agents may produce plausible-sounding outputs that nonetheless contain significant factual fabrications that go undetected by current verification mechanisms. 
Conversely, existing travel planning benchmarks~\cite{xie2024travelplanner, shao2024chinatravel} oversimplify planning into fixed-parameter tool calls over isolated interfaces, thereby abstracting away the core demands of active information seeking and cross-modal alignment.
Existing benchmarks measure whether agents can produce plans that satisfy constraints~\cite{xie2024travelplanner,wang2025tripTailor,singh2024personal} when all relevant facts are provided via structured interfaces. The open question is: 

% \begin{tcolorbox}[width=1.0\textwidth, center, colback=gray!15, colframe=gray!50, boxrule=0.4pt, arc=3pt, left=6pt, right=6pt, top=4pt, bottom=4pt]
%      \textit{Can agents produce factually correct plans when they must actively discover the facts themselves from unstructured evidence?}
% \end{tcolorbox}

\begin{center}
\emph{\textbf{Can agents produce factually correct plans when they must actively discover \\the facts themselves from unstructured evidence?}}
\end{center}

To bridge these gaps, we introduce \textbf{VeriTrip}, a benchmark tailored to evaluate retrieval-based planning capabilities over a reproducible, unstructured, and multimodal web environment. 
VeriTrip establishes a \textbf{Multimodal Retrieval Base (MRB)} containing 8,210 documents and 4,146 images curated from real-world sources.
Rather than querying functions, agents must interact with a frozen snapshot of the web. 
Furthermore, we compel agents to use fuzzy social-media photos as visual anchors. 
This design forces them to resolve real-world visual ambiguities to identify target locations, which is essential to measure true multimodal planning capabilities beyond parametric memorization.
Crucially, to advance evaluation science, we construct a Verifiable Knowledge Base (VKB) to support a rigorous assessment protocol. 
The VKB contains structured facts extracted directly from the MRB and is kept strictly inaccessible to the agents. 
Moving beyond simple format checks, VeriTrip introduces a cell-wise factual verification protocol. 
We automatically cross-reference every generated detail—from transportation ID to accommodation name—against the VKB. 
This source-grounded verification heavily penalizes fabricated facts and effectively decouples genuine logical reasoning from internal parametric hallucinations.
Section~\ref{sec:travelbench} details the construction of these resources, which employs a dual-stage expert-in-the-loop protocol to ensure long-term stability and factual authority.

% Crucially, to ensure factual rigor, we construct a \textbf{Verifiable Knowledge Base (VKB)} by extracting structured facts solely from the MRB documents.
% This enables a source-grounded verification protocol: we quantify whether every claim in the generated itinerary—from ticket prices to POI locations—is strictly supported by the retrieved content, effectively distinguishing reasoning errors from hallucinations.
Building upon our environment, we experiment with several state-of-the-art multimodal large language models (MLLMs), including GPT-4o~\cite{gpt4o}, Claude-4.5-Sonnet~\cite{claude4_5}, and Gemini~\cite{comanici2025gemini}.
The results demonstrate that solving complex retrieval-based planning tasks is extremely challenging. 
While models can maintain high delivery rates, they systematically fail to satisfy factual reliability and fine-grained user preferences. 
Through extensive analysis, we identify a critical cognitive load competition: forcing agents to use visual tools resolves ambiguity but exhausts their reasoning capacity, causing a collapse in high-level preference fulfillment. 
Counter-intuitively, we find that this uncertainty-driven retrieval actually improves factual grounding, as it suppresses the models' reliance on hallucination-prone parametric memory. 
These results expose a significant capability gap between ``seeing'' and ``planning'', highlighting the need for further development of robust multimodal agents. 
\section{Related Works}
\label{sec:relatedworkds}
%-------------------------------------------------------------------------
\subsection{Travel Planning Benchmarks}

Recent benchmarks~\cite{chen2024travelagent,shao2024chinatravel} in travel planning have established important foundations for agent evaluation. TravelPlanner\cite{xie2024travelplanner} was seminal in defining a detailed query format and proposing rule-based criteria for validating plans. Building upon this, subsequent works expanded the scope and sophistication of the task. TripTailor\cite{wang2025tripTailor} significantly increased the scale of travel-related data and introduced the now-prominent ``LLM-as-a-judge" methodology for more nuanced assessment. To enhance realism, benchmarks like ChinaTravel\cite{shao2024chinatravel} and TripScore\cite{qu2025tripscore} curated large datasets of authentic user requests. 
However, a critical limitation unifies these works: they all confine the planning agent to a sanitized API environment.
This paradigm, while valuable for assessing logical reasoning over clean data, fundamentally abstracts away the challenge of autonomous information seeking, allowing agents to bypass the complex retrieval and verification of unstructured, noisy, and ambiguous web content. Our work is proposed to address this gap.

% Prior travel planning benchmarks such as TravelPlanner\cite{xie2024travelplanner} have progressed from defining queries and rule-based criteria to introducing large-scale data, LLM-as-a-judge methods (TripTailor\cite{wang2025tripTailor}), and authentic user requests such as  ChinaTravel\cite{shao2024chinatravel} and TripScore\cite{qu2025tripscore}. However, a critical limitation unifies these approaches: they all confine the agent to sanitized, structured data from APIs. This focus on clean data overlooks the challenge of synthesizing information from the unstructured and ambiguous sources inherent in real-world planning. Our work is proposed to directly address this gap.

%-------------------------------------------------------------------------

\subsection{Deep Research Benchmarks}
Benchmarks for research-oriented agents have evolved from factoid QA tasks (e.g., GAIA~\cite{mialon2023gaia}, BrowseComp~\cite{wei2025browsecomp}) to complex, open-ended challenges.
Recent works like DeepResearchBench~\cite{du2025deepresearch} and DeepScholarBench~\cite{patel2025deepscholar} require agents to synthesize reports from disparate sources, evaluating outputs against pre-compiled ground truth.
To address the volatility of the live web, BrowseComp-Plus\cite{chen2025browsecompplus} and DeepResearchGym\cite{coelho2025deepresearchgym} introduced closed-world sandbox environments using static web corpora to ensure reproducibility.
Our work adopts this sandbox paradigm, integrating complex planning tasks into a controlled environment.
Crucially, we combine this unstructured retrieval setting with the rigorous, transparent evaluation protocols typical of API-based travel benchmarks, ensuring both ecological validity and scientific reproducibility.

% The evaluation of research-oriented agents has evolved from atomic question-answering (QA) benchmarks like GAIA\cite{mialon2023gaia} and BrowseComp\cite{wei2025browsecomp} towards more ecologically valid, open-ended tasks. Recent works, such as DeepResearchBench\cite{du2025deepresearch,bosse2025deep} and DeepScholarBench\cite{patel2025deepscholar}, require agents to synthesize reports from diverse sources, evaluating them against pre-compiled facts. To ensure reproducible evaluation, a key methodological shift has been towards sandboxed environments that use static web corpora (e.g., BrowseComp-Plus\cite{chen2025browsecompplus}, DeepResearchGym\cite{coelho2025deepresearchgym}). Our work adopts this state-of-the-art sandboxed paradigm, applying it to the distinct challenge of complex, multi-modal planning.

\begin{figure*}[ht]
\centering
\includegraphics[width=1.0\linewidth]{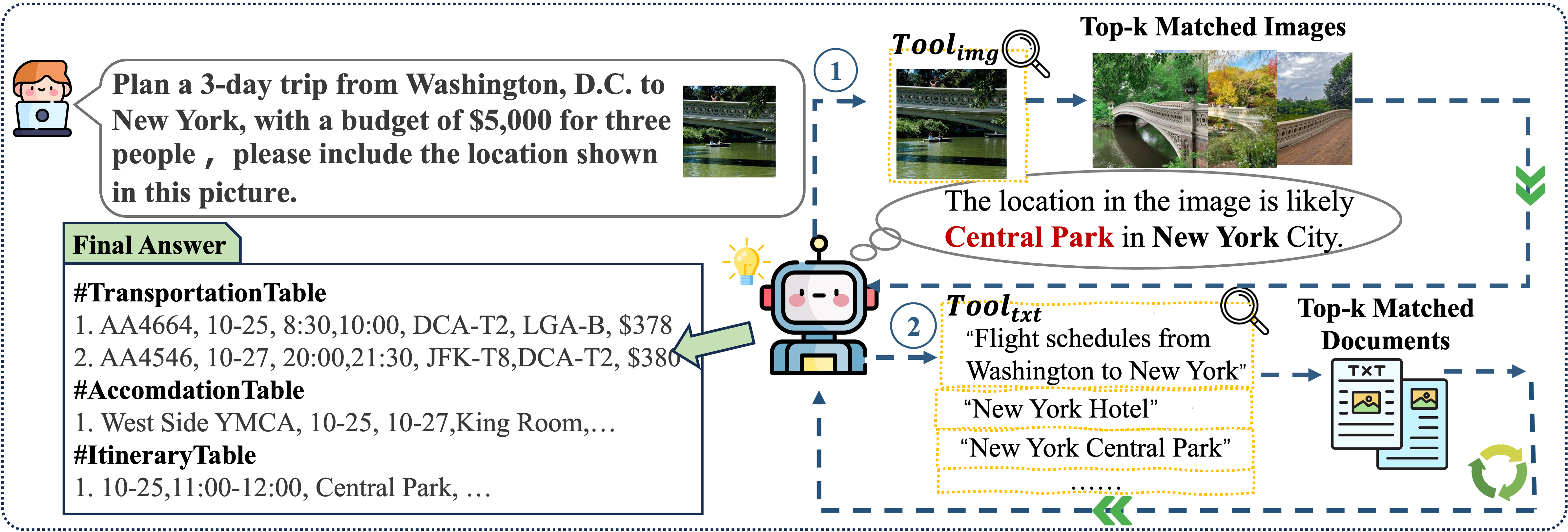} 
\caption{ 
A high-level retrieval-based planning task that can be fully executed in VeriTrip. 
Success requires sophisticated cross-modal disambiguation and active information seeking. 
To accomplish the user's goal, an agent needs to (1) leverage visual tools to resolve ambiguous visual anchors (e.g., social-media photos), (2) iteratively invoke textual tools to search and gather raw evidence from the unstructured Multimodal Retrieval Base (MRB), and (3) synthesize a factually grounded itinerary without relying on parametric memory.}
\label{fig:intro}
\end{figure*}

\section{VeriTrip: An Unstructured Web Environment for Planning Agents}
\label{sec:travelbench}
To rigorously evaluate an agent's ability to synthesize factual itineraries from unstructured web data, we introduce VeriTrip. This section defines the planning task, details the construction of our high-fidelity environment and verifiable annotations, and outlines our source-grounded evaluation protocol along with its interpretive boundaries.

% \subsection{Formulating a Novel Task: Web Multimodal Travel Planning}
\subsection{Task Formulation \& Evaluative Claims}
\label{sub_sec:task_formulation}
We formulate travel planning as an open-retrieval, multimodal task within an unstructured web environment.

\textbf{Input \& Output:} The agent receives a query $Q = (q_{txt}, q_{img})$, where $q_{txt}$ specifies user-level constraints (e.g., duration, budget, dining preferences) and $q_{img}$ provides a visual anchor (e.g., a heavily cropped social-media photo of a target destination).
The expected output is a structured JSON itinerary containing day-by-day plan items and key attributes (e.g., flight numbers, precise POI names, opening hours)

\textbf{Environment:} 
We construct a static evaluation environment called Multimodal Retrieval Base (MRB). 
MRB is a frozen snapshot of web-sourced multimodal documents collected from official tourism websites, municipal announcements, booking pages, and user-posted travel content. 
By restricting agent interactions exclusively to this fixed tool interface, the MRB provides a standardized sandbox for evaluating retrieval and reasoning capabilities without the volatility of the live web.

% \textbf{Agent Interface:}
% The agent follows the ReAct paradigm~\cite{yao2022react} to interleave retrieval actions and generation. It interacts with MRB through a fixed toolset  $\mathcal{T} = \{ T_{txt}, T_{img}, T_{doc},T_{res}\}$:
% \begin{itemize}
%     \item $T_{\text{txt}}(q)$ retrieves relevant text snippets from MRB.
%     \item $T_{\text{img}}(q_{\text{img}})$ performs visual retrieval over MRB using $q_{img}$ and returns the top-$k$ matched images.
%     \item $T_{doc}(doc\_id)$ returns the full content of a document for cross-modal inspection.
%     \item $T_{res}$ use to 
% \end{itemize}
\textbf{Agent Interface \& Action Space.} 
To simulate real-world information seeking, we equip agents with a constrained toolset comprising dual search interfaces and specialized functions. 
For the unstructured MRB, agents utilize a text-based search for document retrieval and an image-based search to resolve visual anchors for entity identification. 
However, we recognize that precise dining metadata is notoriously scarce and fragmented across raw web pages. To address this and ensure a complete evaluation, we augment the environment with specialized functions that return structured restaurant metadata. 
Importantly, agents are still encouraged to actively retrieve user-posted travel guides from the MRB to discover supplementary qualitative information. 
Detailed descriptions of all tools available in VeriTrip are listed in Table~\ref{ap_tab:tool}.

\begin{table*}[ht]
\centering
\small
\caption{Detailed definitions of the agent toolset available in VeriTrip.}
\resizebox{\linewidth}{!}{
\begin{tabular}{l p{9.5cm}}
\toprule
\textbf{Toolset} & \textbf{Description} \\
\midrule
\texttt{docSearch(query: str)}             & Retrieves top-$k$ relevant text snippets and their \texttt{docID}s from the unstructured MRB via keyword search. \\
\midrule
\texttt{imgSearch(imgPath: str)}          & Retrieves top-$k$ visually similar images from the MRB given a reference visual anchor. \\
\midrule
\texttt{getDocument(docID: str)}             & Fetches the full text content of a specific document using its \texttt{docID}. \\
\midrule
\texttt{getRecommendRestaurant(city: str)}             & Lists top recommended restaurants within a specified city. \\
\midrule
\texttt{getRestaurantByName(city: str, name: str)}              & Fetches detailed structured metadata (e.g., pricing, opening hours) for a specific restaurant. \\
\midrule
\texttt{getRestaurantByFood(city: str, food: str)}              & Searches for restaurants offering a specific dish or cuisine in a designated city. \\
\bottomrule
\end{tabular}}
\label{ap_tab:tool}
\end{table*}

\textbf{Evaluative Claims:}
Our primary claim is that true planning capabilities cannot be decoupled from information verification. Unlike function-call-based benchmarks where facts are cleanly provided, VeriTrip claims to test whether an agent can autonomously resolve visual ambiguity and extract factually accurate constraints from noisy, conflicting web evidence without relying on hallucination-prone parametric memory.

\subsection{Environment and Dataset Construction}
\label{sub_sec:corpus}
To rigorously test retrieval-based planning capabilities, we construct a static environment comprising the Multimodal Retrieval Base (MRB) and the Verifiable Knowledge Base (VKB). 
This design is grounded in the assumption that freezing the web into a static sandbox can successfully isolate the volatility of the live web while preserving the complex, unstructured noise of real-world information.
\begin{figure*}[ht]
\centering
\includegraphics[width=1.0\linewidth]{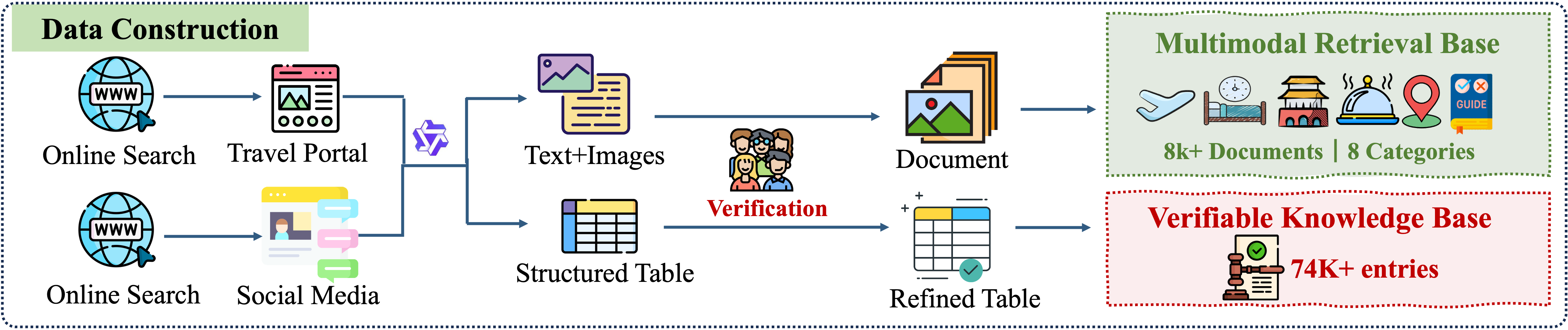} 
\caption{Creating a rigorous retrieval-based benchmark requires balancing authentic web noise with precise factual evaluation. To achieve this, we (1)collect raw multimodal documents from major travel portals and forums, (2)utilize Qwen3-Max as an intelligent semantic filter to prune advertisements while preserving natural noise to form the unstructured MRB, (3)employ Qwen3-Max as a structural parser to extract key entities, followed by strict human cross-checking, to construct VKB.}
\label{fig:data}
\end{figure*}

\textbf{MRB: The High-Fidelity Sandbox.}
The MRB serves as the sole information source accessible to the agents. 
We curated 8,210 multimodal documents covering 15 prominent tourist cities across China and the United States.
To make the construction process cost-efficient and accurate, we employ an auto MRB construct pipeline, as shown in Figure~\ref{fig:data}. 
Specifically, we first deploy automated web scrapers to collect raw data from major travel portals (e.g., Trip.com) and online forums, capturing diverse information including attractions, hotels, transportation schedules, and user reviews.
To ensure corpus quality while preserving authentic web noise, we utilize Qwen3-Max~\cite{qwen3-vl-max} as an intelligent semantic filter to assist human annotators in data sanitization. 
While traditional rule-based scripts handle basic HTML stripping, Qwen3-Max is deployed to comprehend the context and accurately prune stealthy advertisements, promotional spam, and Personally Identifiable Information (PII) embedded within user reviews.
Crucially, as a deliberate design choice, Qwen3-Max is strictly prompted to preserve natural user subjectivity—such as outdated opinions, conflicting reviews, or factual errors—thereby simulating the realistic, noisy environment of the actual internet.

\textbf{VKB: The Verifiable Gold Standard.}
To address the lack of reliable factual evaluation in previous works, we construct the Verifiable Knowledge Base (VKB) by rigorously extracting structured key attributes from MRB documents. 
Inaccessible to agents, VKB acts as the ground truth for an automated cell-wise verification protocol, enabling us to precisely determine whether an agent's plan details are factually grounded in the provided documents or fabricated.
We use Qwen3-Max~\cite{yang2025qwen3} solely as a parser to locate and extract information from the unstructured MRB into our schemas.
Crucially, human annotators then manually cross-check every entry against the raw documents with manageable annotation costs.
Geographic coordinates are exported via the AMap API~\cite{amap_api} and remain essentially unchanged over time.

\begin{wraptable}[15]{r}{0.6\linewidth}  % ← 行数建议从12调至14，防止文字挤压
    \centering
    % 左侧：图片垂直居中 + 限制高度
    \begin{minipage}[c]{0.45\linewidth}
        \centering
        \includegraphics[width=\linewidth, height=3.8cm, keepaspectratio]{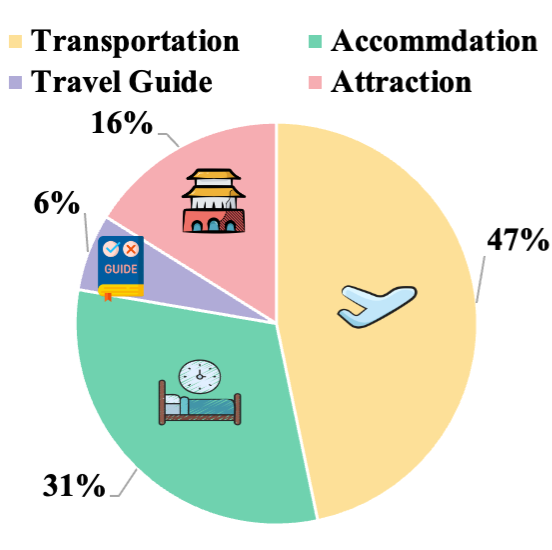}
    \end{minipage}%
    \hfill  % 自动撑满水平间距，比固定 \hspace 更稳定
    % 右侧：表格垂直居中 + 宽度适配minipage
    \begin{minipage}[c]{0.48\linewidth}
        \scriptsize
        \setlength{\tabcolsep}{2pt}
        \resizebox{\linewidth}{!}{%
        \begin{tabular}{@{}lc@{}}
            \toprule
            \textbf{VKB} & \textbf{Data Entries} \\
            \midrule
            Transportation & 57,773 \\
            Restaurants    & 4,304  \\
            Accommodations & 3,613  \\
            Attractions    & 1,056  \\
            POIs           & 7,400  \\
            \bottomrule
        \end{tabular}}
        \vspace{0.6em}  % 两表间距
        \resizebox{\linewidth}{!}{%
        \begin{tabular}{@{}lcccc@{}}
            \toprule
            \textbf{Diff.} & \textbf{Days} & \textbf{Travelers} & \textbf{Pref.} & \textbf{Count}\\
            \midrule
            Simple  & 2--5 & 1--2 & 1--4  & 78 \\
            Medium  & 1--6 & 2--5 & 2--5  & 76 \\
            Complex & 3--7 & 2--8 & 2--8  & 74 \\
            \bottomrule
        \end{tabular}}
    \end{minipage}
    \caption{Overall dataset statistics of VeriTrip.}
    \label{tab:stats_data}
\end{wraptable}
\textbf{Query Construction.} The query set is designed to evaluate an agent's ability to synthesize information under diverse constraints. We adopt a three-stage generation process: First, we utilize Qwen3-Max to create queries from real user-posted travel guides. Second, we select an ambiguous photo as a visual anchor, ensuring that the target point of interest (POI) shown in $q_{img}$ appears in MRB with sufficient but non-trivial evidence. Third, these instances are enriched with a spectrum of user preferences (e.g., budget, dining interests) to create varied difficulty levels. VeriTrip contains 8 types of topics: transportation, restaurants, accommodations, attractions, confusing attractions (out-of-date/false information), travel guides, POIs, and user comments. We categorize the queries into three difficulty levels: simple, medium, and complex. 
As shown in Table~\ref{tab:stats_data}, the distribution is balanced across these levels to ensure robust evaluation.

\subsection{Quality Control}
To ensure a fair and reliable benchmark, we implement a rigorous three-phase validation protocol:

1) \textbf{Temporal Alignment: }
We construct MRB and VKB from the same synchronized web snapshots. 
All documents used to build MRB and facts recorded in VKB are collected within the same time window and subsequently frozen. 
This protocol eliminates inconsistencies caused by live web updates and prevents queries from becoming unsolvable due to temporal drift.

2)  \textbf{Human Review: }
Each query undergoes manual inspection to verify that at least one feasible solution exists under the given constraints using only the provided MRB documents and tool interface. 
Queries that rely on external knowledge or lack sufficient evidence in the corpus are discarded.

3) \textbf{Empirical Validation:}
We deploy an agent powered by GPT-4.5-preview~\cite{Gpt45} to attempt each query using the standard toolset. 
This step serves exclusively to identify and filter out instances that are empirically unsolvable within our corpus. Note that the plans generated during this phase are neither released nor used as reference answers for evaluation.

4) \textbf{Addressing Evaluator Fairness:}
Addressing Evaluator Fairness: To address concerns that using Qwen3-Max as the initial VKB parser might bias the ground truth, we sampled 1,000 documents and applied GPT-4o and Claude-3.5-Sonnet to extract the same fields. Using Exact Match (EM) for numerical fields and Cosine Similarity (CS) for textual fields, both models yielded $\geq$ 96.8\% EM and $\geq$ 94.7\% CS agreement with the VKB. This confirms the VKB's factual integrity is model-agnostic and anchored strictly in the corpus.

\subsection{Evaluation Protocol}
\label{sec:eval_protocol}

To accurately measure an agent's planning capability within a noisy, unstructured web environment, VeriTrip eschews traditional textual surface-form matching. Instead, we employ a multi-stage, programmatic "functional correctness" evaluation pipeline. When an agent submits its final itinerary as a structured JSON, we conduct the following automated verification steps. A high-level mapping of these dimensions to their underlying programmatic implementations is illustrated in Table~\ref{tab:eval_implementation}.

\begin{itemize}[leftmargin=*,nosep]
    \item \textbf{Format Check } The initial stage ensures the output is a strictly parsable JSON. The proportion of queries that successfully pass this structural integrity check defines the \emph{Delivery Rate (DR)}. Invalid outputs receive no further assessment.
    
    \item \textbf{Cell-wise Fact Check} This stage detects whether the agent's output is strictly grounded in the MRB or fabricated from parametric memory. We disaggregate the generated plan into fine-grained entities (e.g., transportation IDs, departure timestamps, hotel names) and cross-reference them against the VKB. We enforce \texttt{exact\_match} for rigid numerical and temporal data, and utilize \texttt{fuzzy\_match} (via semantic similarity) for textual entities. The percentage of absolutely correct factual cells defines the \emph{Factual Reliability (FR)}.
    
    \item \textbf{Commonsense \& Preference Check:} We deploy a suite of deterministic programmatic evaluators to assess whether the final plan state satisfies user constraints. Hard constraints (e.g., budget limits, chronological continuity) are measured by the \emph{Pass Rate (PR)} at both the micro and macro levels. Soft constraints (e.g., dining tastes, accommodation types) are measured by the \emph{Preference Fulfillment Rate (PFR)}, representing the proportion of plans that satisfy nuanced user profiles.
    
    \item \textbf{Geographic Coherence Check:} To evaluate spatial common sense, we extract all unique Points of Interest (POIs) from the generated plan and map them to real-world coordinates. We then compare the agent's total travel distance against a theoretical optimal distance calculated by a Traveling Salesperson Problem (TSP) solver. The normalized difference yields the \emph{Average Margin (AM)}, representing the excess travel distance per POI.
\end{itemize}

\begin{table*}[ht]
\centering
\small
\caption{The programmatic evaluation implementation of VeriTrip. We translate complex constraints into executable metrics, ensuring a transparent and reproducible functional correctness assessment.}
\resizebox{\linewidth}{!}{
\begin{tabular}{l l l}
\toprule
\textbf{Evaluation Dimension} & \textbf{Metric} & \textbf{Eval Implementation / Logic} \\
\midrule
\textbf{Format Check} & \multirow{2}{*}{DR $\uparrow$} & \multirow{2}{*}{\texttt{parsed\_plan = parse\_json(agent\_output)}} \\
\textit{(Is the output a strictly valid JSON?)} & & \\
\midrule
\textbf{Fact Check: Transportation} & \multirow{2}{*}{FR $\uparrow$} & \texttt{time = parse\_time(agent\_time)} \\
\textit{(Does the flight exist?)} & & \texttt{exact\_match(time, VKB.transportation.time)} \\
\midrule
\textbf{Fact Check: POI Entities} & \multirow{2}{*}{FR $\uparrow$} & \multirow{2}{*}{\texttt{fuzzy\_match(agent\_POI\_name, VKB.attractions.name)}} \\
\textit{(Is the attraction real?)} & & \\
\midrule
\textbf{Commonsense: Budget} & \multirow{2}{*}{PR$_{mi}$ $\uparrow$} & \texttt{total\_cost = sum(transport, hotel, tickets)} \\
\textit{(Does it exceed \$2000?)} & & \texttt{is\_within\_budget = total\_cost <= 2000*1.5} \\
\midrule
\textbf{Preference: Dining} & \multirow{2}{*}{PFR $\uparrow$} & \multirow{2}{*}{\texttt{must\_include(agent\_meals, user\_meal\_preference)}} \\
\textit{(Did they eat local food?)} & & \\
\midrule
\textbf{Geographic Coherence} & \multirow{2}{*}{AM $\downarrow$} & \texttt{D\_ref = TSP\_Solver(agent\_POIs\_coordinates)} \\
\textit{(Is the route spatially logical?)}& & \texttt{AM = max(0, D\_agent - D\_ref) / len(POIs)} \\
\bottomrule
\end{tabular}
}
\label{tab:eval_implementation}
\end{table*}
\vspace{-2mm}
% \textbf{Evaluation Design Rationale \& Interpretive Boundaries.} 
% Our evaluation is designed to programmatically assess "Functional Correctness" rather than sequence-level similarity. The rationale behind the \textbf{Cell-wise Fact Check} is to strictly penalize fabrications. If a model generates a correct real-world hotel price that is absent from the frozen MRB, it is penalized. This design forces the decoupling of \textit{genuine logical reasoning} from \textit{internal parametric hallucinations}. Furthermore, the rationale for using a \textbf{TSP Solver} is to provide an objective mathematical proxy for geographic common sense, assessing whether the model can cluster activities rationally without predefined routing APIs.

% \textbf{Interpretive Boundaries.} Following rigorous benchmarking principles, we explicitly state the boundaries of our evaluative claims. VeriTrip evaluates an agent's capacity for static evidence retrieval, cross-modal disambiguation, and multi-constraint planning within a frozen, evidence-sufficient sandbox. Consequently, our findings explain how well models synthesize known, unstructured facts. They \textbf{cannot and should not be infinitely extrapolated to real-time, dynamic booking environments} (e.g., executing live ticket purchases or responding to CAPTCHAs and dynamic pricing), which require distinct interactive execution capabilities.

\textbf{Evaluation Design Rationale \& Interpretive Boundaries.} 
Our evaluation is designed to programmatically assess "Functional Correctness". The rationale behind the Cell-wise Fact Check is to strictly penalize fabrications. 
This design forces the decoupling of genuine logical reasoning from internal parametric hallucinations. 
Furthermore, the rationale for using a TSP Solver is to provide an objective mathematical proxy for "geographic common sense", assessing if the model can cluster activities rationally without relying on predefined API routing tools.

\textbf{Interpretive Boundaries}: Following rigorous benchmarking principles, we explicitly state the boundaries of our evaluative claims. 
VeriTrip evaluates an agent’s capacity for static evidence retrieval, cross-modal disambiguation, and multi-constraint planning within a frozen, evidence-sufficient sandbox. 
Consequently, our findings explain how well models synthesize known, unstructured facts. 
They cannot and should not be infinitely extrapolated to real-time, dynamic booking environments (e.g., executing live ticket purchases or responding to real-time sold-out scenarios), which require distinct interactive execution capabilities.
\definecolor{lightblue}{RGB}{220,235,250}
\definecolor{darkblue}{RGB}{75,137,172}
\tcbset{
  takeawaysbox/.style={
    title=Takeaways,
    colback=lightblue!80,
    colframe=darkblue,
    fonttitle=\bfseries\small,
    coltitle=white,
    colbacktitle=darkblue,
    enhanced,
    attach boxed title to top left={xshift=2.5mm,yshift=-2.5mm},
    boxed title style={rounded corners, size=small, colframe=darkblue, colback=darkblue},
    width=\linewidth,
    arc=3.5mm
  }
}

\section{Experiments}

\subsection{Setup}
% \textbf{Baselines: VLMs Search Agents}
Our comprehensive evaluation encompasses a diverse range of state-of-the-art models with strong agentic search capabilities, including both proprietary and open-source MLLMs.
The models are categorized into two groups: 

\textbf{MLLMs without Thinking:}GPT-5-mini~\cite{gpt5}, GPT-4o~\cite{hurst2024gpt}, GPT-4o-mini~\cite{gpt4omini}, Gemini-2.5-flash~\cite{comanici2025gemini},  Claude-4.5-Sonnet~\cite{claude4_5}, Claude-3.7-Sonnet~\cite{claude3_7}, Qwen-VL-Max~\cite{qwen3-vl-max}, Qwen3-VL-235B~\cite{yang2025qwen3}.
\textbf{MLLMs with Thinking:} o3~\cite{gpto3}, o4-mini~\cite{gpto4mini}, Gemini-2.5-pro~\cite{comanici2025gemini}.

We use Qwen3-Embedding-8b~\cite{zhang2025qwen3} as $T_{txt}$, and serve it with Tevatron~\cite{ma2025tevatron} dense retrieval toolkit.
DINOv2~\cite{oquab2023dinov2} as $T_{img}$ for image retrieval, which utilizes ViT-S/14 as its backbone.
Please refer to the Appendix~\ref{ap:exp_detail} for detailed experiment setups.

% \textbf{Metrics.} To quantify the agent's performance across the different stages of our evaluation protocol, we define five metrics aligned with the capability taxonomy: \textbf{DR}: proportion of structurally parsable plans; \textbf{FR}: proportion of verifiable factual cells matching VKB. \textbf{PR$_{mi}$}/\textbf{PR$_{ma}$}: constraint adherence at micro/macro level; \textbf{PFR}: proportion of plans satisfying all preferences; \textbf{AM}: spatial inefficiency vs. TSP lower bound (auxiliary). Full definitions and formulas are in Appendix. 

\subsection{Main Results}
Table~\ref{exp:vlms} reveals a systemic failure across all tested MLLMs when confronted with unconstrained retrieval-based planning. 
Due to space limitations, we only report on simple and complex tasks; the complete tables are reported in the Appendix.
We have the following observations:

\noindent \textbf{Correlation between Active Retrieval and Factual Reliability.} 
Table 2 reveals a critical link between agent behavior and grounding accuracy. 
First, we observe a distinct ``laziness" on simple tasks: almost all models exhibit lower tool usage and correspondingly lower FR scores on simple queries compared to complex ones (e.g., o4-mini improves FR by 8.74\% when task complexity forces more tool interactions). This confirms that forcing agents out of their parametric memory into active retrieval is key to reducing hallucinations.
Second, high-performing models like Claude-4.5-Sonnet demonstrate a ``Deep Research" pattern, averaging nearly 40 tool calls per task—triple that of GPT-4o. This extensive information seeking directly correlates with its state-of-the-art FR performance (68.6\%), suggesting that widely-used API-centric agents (like GPT-4o) may be ``under-searching" in unstructured environments.
However, retrieval volume is not a panacea; Qwen3-VL-235B achieves high tool calls but strictly average FR (45.57\%), indicating a gap in information extraction efficiency—the model retrieves documents but struggles to accurately transcribe cell-level facts from the noisy web context.

\begin{table*}[ht]
\caption{Main results of different MLLMs on VeriTrip. AM is measured in units of 10 km. Tool Calls indicates the number of tool calls. 
The colored values indicate the performance delta (Complex score - Simple score) for each metric. \textcolor{red}{Red} denotes an improvement, while \textcolor{ForestGreen}{green} denotes a degradation. All subsequent experiments followed this setting.}

\resizebox{\textwidth}{!}{

\begin{tabular}{l ccccccc ccccccc}
\toprule
 \multirow{2}{*}{\textbf{Model}}                  & \multicolumn{6}{c}{Simple}                                      & \multicolumn{6}{c}{Complex}               \\ \cmidrule(lr){2-7} \cmidrule(lr){8-14}   
              & DR$\uparrow$ & FR$\uparrow$  & PFR$\uparrow$ & PR$_{mi}$$\uparrow$    & AM$\downarrow$ & TC & DR$\uparrow$ & FR$\uparrow$ & PFR$\uparrow$ & PR$_{mi}$$\uparrow$ & AM$\downarrow$ &TC   \\ \midrule

\multicolumn{14}{c}{\textbf{MLLMs without thinking}} \\
\midrule
Claude-4.5-Sonnet
& 69.23 & 62.81 & 64.10 & 89.42 & 8.49  & 32  & 77.03 {\scriptsize \color{red}{+7.80}} & 68.60 {\scriptsize \color{red}{+5.79}} & 50.00 {\scriptsize \color{ForestGreen}{-14.10}} & 85.84 {\scriptsize \color{ForestGreen}{-3.58}} & 16.59 {\scriptsize \color{ForestGreen}{+8.10}} & 37   \\
Claude-3.7-Sonnet 
& 66.16 & 63.81 & 58.12 & 82.17 & 10.97 & 6   & 68.33 {\scriptsize \color{red}{+2.17}} & 65.89 {\scriptsize \color{red}{+2.08}} & 44.59 {\scriptsize \color{ForestGreen}{-13.53}} & 79.25 {\scriptsize \color{ForestGreen}{-2.92}} & 14.94 {\scriptsize \color{ForestGreen}{+3.97}} & 8   \\
GPT-5-mini 
& 67.26 & 42.02 & 52.62 & 87.68 & 7.18  & 8   & 71.67 {\scriptsize \color{red}{+4.41}} & 49.06 {\scriptsize \color{red}{+7.04}} & 34.06 {\scriptsize \color{ForestGreen}{-18.56}} & 80.43 {\scriptsize \color{ForestGreen}{-7.25}} & 8.82 {\scriptsize \color{ForestGreen}{+1.64}}  & 12  \\
GPT-4o             
& 63.64 & 48.21 & 45.20 & 88.02 & 3.12  & 8   & 69.57 {\scriptsize \color{red}{+5.93}} & 51.44 {\scriptsize \color{red}{+3.23}} & 32.53 {\scriptsize \color{ForestGreen}{-12.67}} & 80.12 {\scriptsize \color{ForestGreen}{-7.90}} & 7.34 {\scriptsize \color{ForestGreen}{+4.22}}  & 11  \\
GPT-4o-mini      
& 57.69 & 41.88 & 28.21 & 79.21 & 10.05 & 7   & 50.00 {\scriptsize \color{ForestGreen}{-7.69}} & 53.15 {\scriptsize \color{red}{+11.27}} & 6.76 {\scriptsize \color{ForestGreen}{-21.45}}  & 78.91 {\scriptsize \color{ForestGreen}{-0.30}} & 4.91 {\scriptsize \color{red}{-5.14}}  & 10 \\
Gemini-2.5-flash 
 & 55.17 & 44.11 & 45.93 & 75.39 & 12.81 & 7   & 49.24 {\scriptsize \color{ForestGreen}{-5.93}} & 47.85 {\scriptsize \color{red}{+3.74}} & 36.21 {\scriptsize \color{ForestGreen}{-9.72}} & 73.80 {\scriptsize \color{ForestGreen}{-1.59}} & 20.17 {\scriptsize \color{ForestGreen}{+7.36}} & 12 \\

Qwen-VL-Max    
 & 61.90 & 42.09 & 64.61 & 84.48 & 10.55 & 21  & 68.52 {\scriptsize \color{red}{+6.62}} & 45.56 {\scriptsize \color{red}{+3.47}} & 46.22 {\scriptsize \color{ForestGreen}{-18.39}} & 76.84 {\scriptsize \color{ForestGreen}{-7.64}} & 10.01{\scriptsize \color{red}{-0.54}} & 24  \\
Qwen3-VL-235B     
& 68.26 & 41.34 & 66.46 & 84.28 & 4.84  & 19  & 64.30 {\scriptsize \color{ForestGreen}{-3.96}} & 45.57 {\scriptsize \color{red}{+4.23}} & 36.49 {\scriptsize \color{ForestGreen}{-29.97}} & 79.10 {\scriptsize \color{ForestGreen}{-5.18}} & 9.58 {\scriptsize \color{ForestGreen}{+4.74}}  & 29  \\
\midrule
\multicolumn{15}{c}{\textbf{MLLMs with thinking}} \\
\midrule
o3  & 65.38 & 46.81 & 55.38 & 87.76 & 8.93  & 25  & 65.63 {\scriptsize \color{red}{+0.25}} & 53.43 {\scriptsize \color{red}{+6.62}} & 27.57 {\scriptsize \color{ForestGreen}{-27.81}} & 82.51 {\scriptsize \color{ForestGreen}{-5.25}} & 10.04 {\scriptsize \color{ForestGreen}{+1.11}} & 31   \\

o4-mini 
& 64.69 & 40.21 & 50.77 & 85.50 & 6.43  & 12  & 63.23 {\scriptsize \color{ForestGreen}{-1.46}} & 48.95 {\scriptsize \color{red}{+8.74}} & 21.89 {\scriptsize \color{ForestGreen}{-28.88}} & 76.23 {\scriptsize \color{ForestGreen}{-9.27}} & 11.92 {\scriptsize \color{ForestGreen}{+5.49}} & 19   \\

Gemini-2.5-pro  
 & 60.15 & 41.83 & 48.5  & 77.63 & 11.97 & 11  & 69.51 {\scriptsize \color{red}{+9.36}} & 44.88 {\scriptsize \color{red}{+3.05}} & 22.37 {\scriptsize \color{ForestGreen}{-26.13}} & 75.38 {\scriptsize \color{ForestGreen}{-2.25}} & 15.95 {\scriptsize \color{ForestGreen}{+3.98}} & 14  \\
\midrule
\bottomrule
\end{tabular}

}
\label{exp:vlms}
\end{table*}

\noindent \textbf{Gap between Logical Reasoning and Intent Fulfillment.}
While task complexity boosts FR by compelling agents to retrieve evidence, it cripples adherence to the plan's holistic conditions, revealing a fundamental trade-off. As shown in Table 2, this is most starkly illustrated by the collapse of the PFR in complex scenarios, where scores plummet across all models (e.g., GPT-4o-mini: -21.45\%, o3: -27.81\%).
We identify a conflict between the low-level task of verifying atomic facts (for FR) and the high-level task of satisfying a system of conditions (for PFR \& PR). The cognitive load of fact verification appears to exhaust the agent's capacity for the latter. This is especially true for ``thinking" models, where structured reasoning leads to ``process over-fixation" on fact-checking, causing them to neglect global constraints. This reveals a core capability gap: agents excel at grounding individual facts, leaving insufficient capacity to verify if that location actually matches the user's specific attributes, leading to a plan that is factually true but preferentially wrong.

\subsection{How Visual Grounding Affects Planning}
\textbf{Setting.} 
We conduct an ablation study across three configurations to isolate the impact of visual grounding on planning performance: $(q_{txt},q_{img})$ without $T_{img}$, $(q_{txt},q_{img})$ with $T_{img}$ and only $q_{txt}$. We use gold-standard POI names to replace $T_{img}$. 
For the ablation experiments, we use the overall results from VeriTrip. 
We exclude the AM metric from this ablation study as its strong dependence on POI identification prevents a uniform analysis.

\begin{wraptable}{r}{0.53\linewidth} 
\centering
% \vspace{-5mm}
\caption{Ablation study on visual grounding.}
\vspace{-2mm}
\resizebox{\linewidth}{!}{
\begin{tabular}{l ccccc}
\toprule
 \textbf{Setting} & DR $\uparrow$ & FR $\uparrow$ & PFR $\uparrow$ & {PR$_{mi}$} $\uparrow$ & {PR$_{ma}$} $\uparrow$ \\ \midrule
\multicolumn{6}{l}{\textbf{GPT-4o-mini}} \\
\quad (1) $(q_{txt}, q_{img})$ w/o $T_{img}$ & 42.98 & 43.68 & 15.79 & 70.89 & 0.88 \\
\quad (2) $(q_{txt}, q_{img})$ w $T_{img}$  & 49.12 & 47.32 & 17.54 & 73.49 & 1.42 \\
\rowcolor{gray!10}
\quad \textit{$T_{img}$ Gain (2-1)} & \textcolor{red}{+6.14} & \textcolor{red}{+3.64} & \textcolor{red}{+1.75} & \textcolor{red}{+2.60} & \textcolor{red}{+0.54} \\
\cmidrule{1-6}
\quad (3) $q_{txt}$ only (Gold) & 70.43 & 47.94 & 46.22 & 75.05 & 9.46 \\
\rowcolor{orange!5}
\quad \textit{Reality Gap (3-2)} & \textcolor{red}{+21.31} & \textcolor{red}{+0.62} & \textcolor{red}{+28.68} & \textcolor{red}{+1.56} & \textcolor{red}{+8.04} \\
\midrule

% --- GPT-4o Group ---
\multicolumn{6}{l}{\textbf{GPT-4o}} \\
\quad (1) $(q_{txt}, q_{img})$ w/o $T_{img}$ & 68.16 & 48.79 & 43.86 & 79.12 & 1.25 \\
\quad (2) $(q_{txt}, q_{img})$ w $T_{img}$  & 73.45 & 56.12 & 42.50 & 84.12 & 5.56 \\
\rowcolor{gray!10}
\quad \textit{$T_{img}$ Gain (2-1)} & \textcolor{red}{+5.29} & \textcolor{red}{+8.33} & \textcolor{ForestGreen}{-1.36} & \textcolor{red}{+5.00} & \textcolor{red}{+4.31} \\
\cmidrule{1-6}
\quad (3) $q_{txt}$ only (Gold) & 80.26 & 52.60 & 63.16 & 85.48 & 13.16 \\
\rowcolor{orange!5}
\quad \textit{Reality Gap (3-2)} & \textcolor{red}{+6.81} & \textcolor{ForestGreen}{-3.52} & \textcolor{red}{+20.66} & \textcolor{red}{+1.36} & \textcolor{red}{+7.60} \\
\midrule

% --- Qwen Group ---
\multicolumn{6}{l}{\textbf{Qwen-Max-VL}} \\
\quad (1) $(q_{txt}, q_{img})$ w/o $T_{img}$ & 69.08 & 49.81 & 49.46 & 81.04 & 10.83 \\
\quad (2) $(q_{txt}, q_{img})$ w $T_{img}$  & 72.20 & 50.35 & 55.40 & 76.88 & 14.12 \\
\rowcolor{gray!10}
\quad \textit{$T_{img}$ Gain (2-1)} & \textcolor{red}{+3.12} & \textcolor{red}{+0.54} & \textcolor{red}{+5.94} & \textcolor{ForestGreen}{-4.16} & \textcolor{red}{+3.29} \\
\cmidrule{1-6}
\quad (3) $q_{txt}$ only (Gold) & 77.19 & 46.55 & 74.56 & 86.61 & 32.89 \\
\rowcolor{orange!5}
\quad \textit{Reality Gap (3-2)} & \textcolor{red}{+4.99} & \textcolor{ForestGreen}{-3.80} & \textcolor{red}{+19.16} & \textcolor{red}{+9.73} & \textcolor{red}{+18.77} \\
\bottomrule
\end{tabular}

}
\label{tab:img_effect}
\end{wraptable}
\textbf{Analysis.} As shown in Table~\ref{tab:img_effect}, while providing the image search tool ($T_{img}$)  improves the DR by resolving ambiguity, it reveals a critical trade-off rooted in cognitive resource competition.
The cognitive load from visual grounding in 
Setting-2 compromises higher-order planning, causing a sharp PFR decline compared to the ideal Setting-3. 
For example, GPT-4o-mini drops from 46.22$\%$ to 17.54$\%$. 
Counter-intuitively, this same load benefits FR. 
For models like GPT-4o, FR is higher in Setting-2 (56.12$\%$) than in Setting-3 (52.60$\%$), as the inherent uncertainty forces retrieval-based verification, while explicit POI names (Setting-3) may trigger hallucination from parametric knowledge.
In essence, MLLMs struggle to manage the cognitive load from multimodal inputs. 
The resulting uncertainty improves factual grounding by forcing retrieval, but degrades planning quality by consuming reasoning resources.

\subsection{How Noisy Information Affects Planning}
\textbf{Setting.}To examine how agents make decisions when facing conflicting evidence in unstructured web content, we design a controlled ablation on the presence of misleading attraction information in MRB. 
MRB-Clean setting removes misleading documents while keeping the overall corpus size comparable by adding neutral documents that do not introduce additional conflicts.

\textbf{Analysis.} 
As shown in Table~\ref{tab:falase_info}, the experimental results indicate that noisy information generally hinders an agent's ability to correctly identify and retrieve information. 
\begin{wraptable}{r}{0.6\linewidth} 
\centering
\caption{Impact of Noisy Information on model performance.}
\resizebox{\linewidth}{!}{
\begin{tabular}{l ccccccc}
\toprule
\textbf{Setting} & {DR} $\uparrow$ & {FR} $\uparrow$ & {PFR} $\uparrow$ & {PR$_{mi}$} $\uparrow$ & {PR$_{ma}$} $\uparrow$ & {AM} $\downarrow$ \\
\midrule

% --- GPT-4o-mini Group ---
\multicolumn{7}{l}{\textbf{GPT-4o-mini}} \\
\quad {(1)MRB-Clean} & 57.69 & 48.92  & 17.11 & 78.15 & 1.32 & 7.11 \\
\quad (2)MRB & 49.12 & 47.32  & 17.54 & 73.49 & 1.42 & 10.63 \\

\rowcolor{gray!10}
\quad{ \textit{Noise Impact ($\Delta$)}} & \color{ForestGreen}{-8.57} & \color{ForestGreen}{-1.60}  & \color{red}{+0.43} & \color{ForestGreen}{-4.66} & \color{red}{+0.10} & \color{ForestGreen}{+3.52} \\
\midrule

% --- GPT-4o Group ---
\multicolumn{7}{l}{\textbf{GPT-4o}} \\
\quad {MRB-Clean} & 74.32 & 56.12  & 48.12 & 87.55 & 6.82 & 4.20\\
\quad  MRB & 73.45 & 52.65  & 42.50 & 84.12 & 4.56 & 3.45 \\

\rowcolor{gray!10}
\quad{ \textit{Noise Impact ($\Delta$)}} & \color{ForestGreen}{-0.87} & \color{ForestGreen}{-3.53}  & \color{ForestGreen}{-5.62} & \color{ForestGreen}{-3.43} & \color{ForestGreen}{-2.26} & \color{red}{-0.75} \\ 
\midrule

% --- Qwen Group ---
\multicolumn{7}{l}{\textbf{Qwen-VL-Max}} \\
\quad {MRB-Clean} & 78.50 & 46.55  & 64.60 & 81.45 & 5.25 & 9.05 \\
\quad MRB & 72.20 & 50.35  & 55.40 & 76.88 & 3.12 & 10.42 \\

\rowcolor{gray!10}
\quad{ \textit{Noise Impact ($\Delta$)}} & \color{ForestGreen}{-6.30} & \color{red}{+3.80}  & \color{ForestGreen}{-9.20} & \color{ForestGreen}{-4.57} & \color{ForestGreen}{-2.13} & \color{ForestGreen}{+1.37}\\ 
\bottomrule
\end{tabular}

}
\label{tab:falase_info}
\vskip -0.1in
\end{wraptable}
A consistent trend across all models is the decline in DR and {PR$_{mi}$} when noisy documents are present.
While GPT-4o shows a slight improvement in AM under noisy conditions, this may suggest that more capable models possess a higher threshold for ignoring logically inconsistent data. However, the overall low precision scores indicate that most agents still struggle to resolve factual contradictions effectively.
To improve agent robustness in MRB with conflicting information, future research should focus on developing explicit conflict detection mechanisms that identify contradictions before the planning stage. 
Incorporating multi-source cross-verification modules could help agents verify key facts to ensure accuracy.

\begin{figure}[h]
\centering
\includegraphics[width=0.6\linewidth]{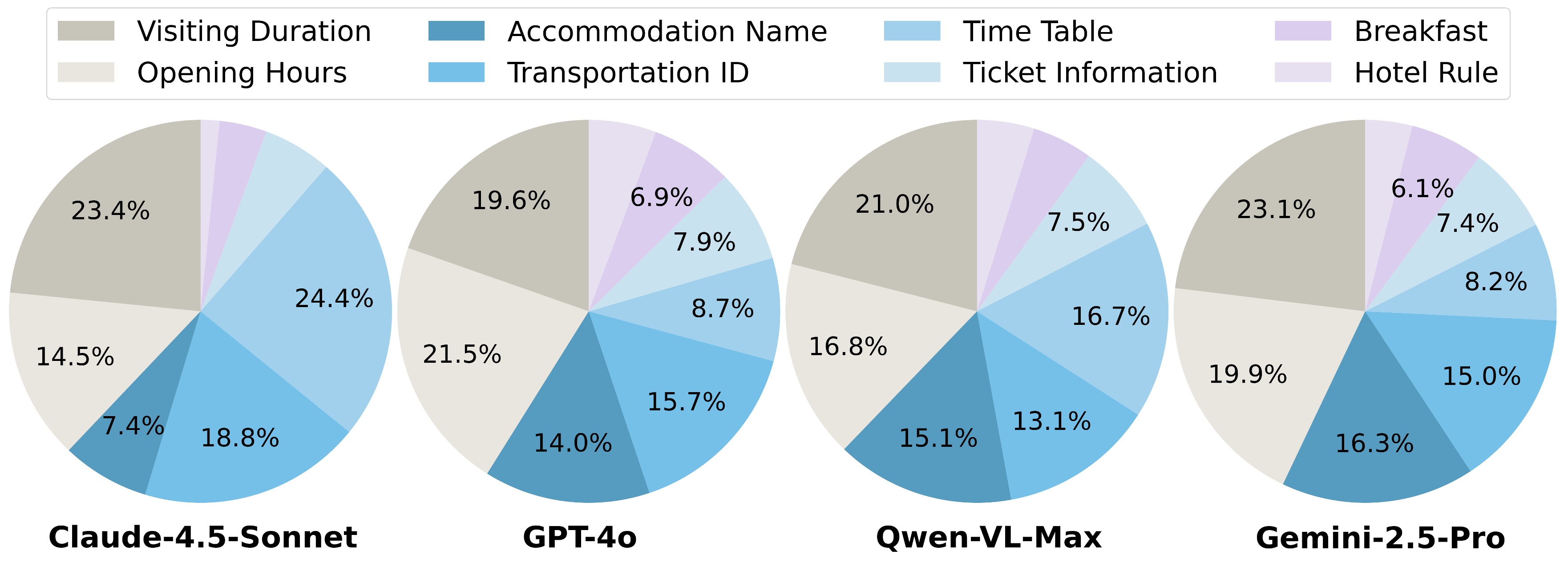} 
\caption{Distribution of factual errors by agents on VeriTrip.} 
\label{fig:fact_error}
\end{figure}

\subsection{Further Analysis}

\textbf{Fact and Constraint Error Analysis}
\label{subsubsec:error_analysis}
An analysis of factual errors, as detailed in Figure~\ref{fig:fact_error}, categorizes agent failures into two primary dimensions. 
First, agents exhibit poor handling of flexible requirements, such as visitation durations, frequently generating schedules that diverge from official recommendations. 
Second, there is a high incidence of inaccuracy regarding rigid factual identifiers, specifically transportation IDs and precise departure/arrival timestamps.
Our investigation reveals a systemic ``heuristic default" flaw: when initial searches fail, agents bypass instructions to re-invoke tools, prematurely relying on internal parametric knowledge and resulting in hallucinations. 
This strategic non-compliance highlights a deficiency in handling retrieval challenges. 
% Furthermore, low micro-pass rates (Fig.~\ref{fig:pass_rate}) across budget, dining, and intra-city commuting reveal a significant bottleneck in complex preference alignment.

\begin{wrapfigure}{r}{0.5\linewidth}
\centering 
\includegraphics[width=\linewidth]{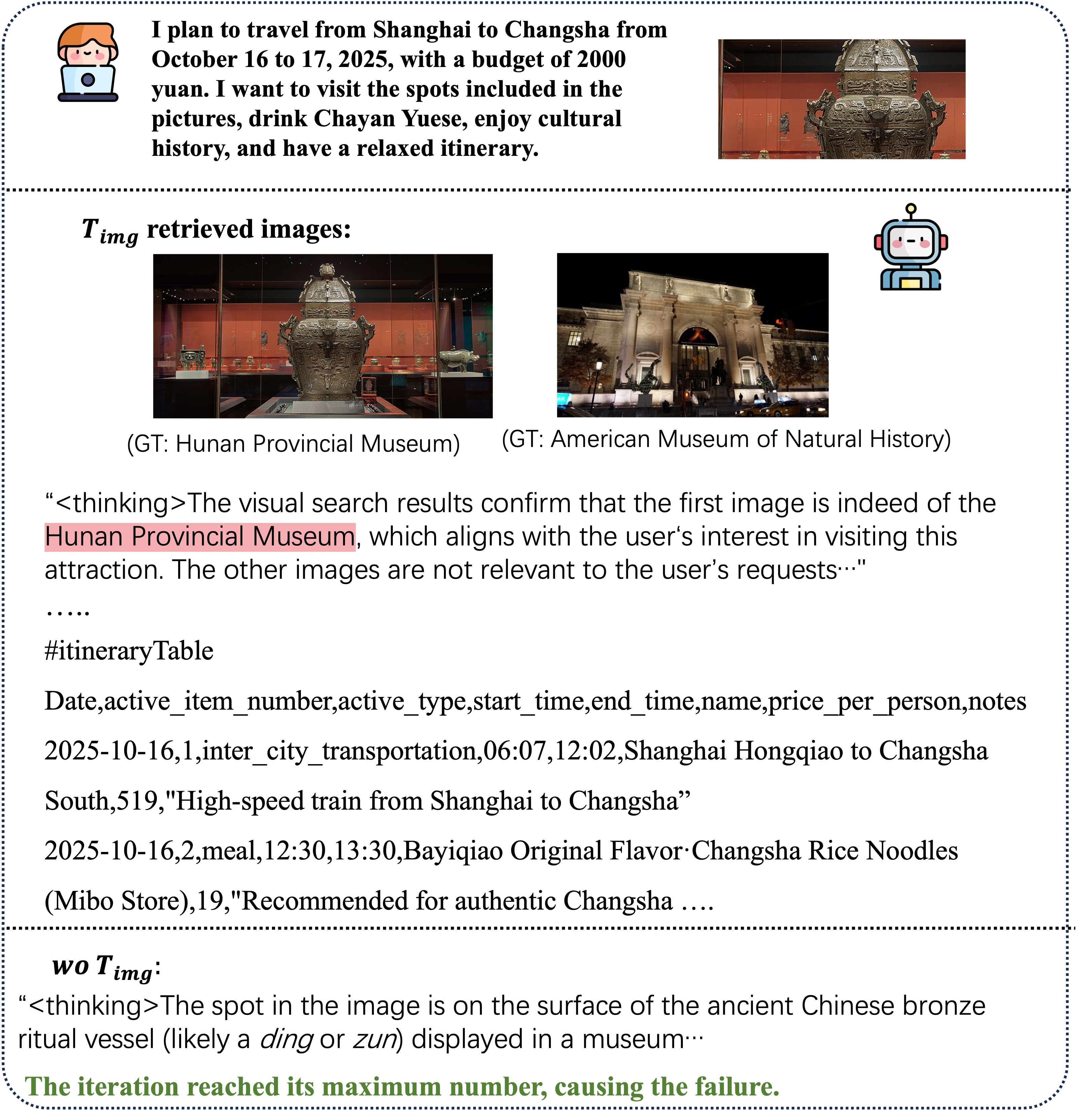}
\caption{Case Study of Qwen-Max-VL. 
}
\label{fig:qwen_max_case}
\end{wrapfigure}
\textbf{Case Study.}
To investigate how visual ambiguity impacts retrieval-based planning, Figure~\ref{fig:qwen_max_case} presents a comparative case study of an agent operating with and without the visual retrieval tool ($T_{img}$). 
The user query provides a heavily cropped image of a bronze artifact and requests a constrained itinerary in Changsha.

\textbf{With Visual Grounding (w/ $T_{img}$):} The agent successfully leverages the visual tool to resolve the ambiguous anchor, aligning the cropped artifact with the POI through cross-modal retrieval results. 
This successful disambiguation provides a concrete textual anchor, allowing the agent's ReAct loop to proceed seamlessly into textual document retrieval and itinerary synthesis.

\textbf{The Perception Collapse (w/o $T_{img}$):} When the visual tool is ablated, forcing the agent to rely entirely on its internal parametric memory, it exhibits a critical perception gap. 
As shown in the bottom panel of Figure~\ref{fig:qwen_max_case}, the agent successfully recognizes the high-level semantics of the image.
However, it fundamentally fails at fine-grained, instance-level recognition to link this artifact to a specific POI in the real world. 
Due to semantic ambiguity, the agent cannot formulate a valid textual search query. 
Consequently, the agent repeatedly loops in the reasoning phase without generating executable tool calls, ultimately triggering the maximum iteration limit. 
This failure highlights our core evaluative claim: current MLLMs lack fine-grained visual discriminative capabilities and critically depend on external retrieval to bridge the fatal gap between ``seeing" and ``planning."

\section{Conclusion and Limitations}

In this work, we bridge the gap between static API-based evaluation and the open-ended reality of autonomous travel agents by introducing VeriTrip. By establishing an unstructured Multimodal Retrieval Base (MRB) and a structured Verifiable Knowledge Base (VKB), we enable the first rigorous, cell-wise factual verification protocol for retrieval-based planning. Our extensive experiments reveal a critical bottleneck in current state-of-the-art MLLMs: a perception-reasoning trade-off. We observe that the heavy cognitive demand of cross-modal disambiguation interferes with instruction following, leading agents to prioritize local grounding accuracy over global preference satisfaction.

\textbf{Limitations.} VeriTrip has several boundaries that point to future research directions:
\begin{itemize}[leftmargin=*,nosep]
    \item \textit{Geographic and Cultural Scope:} The current MRB focuses on 15 prominent tourist cities in China and the United States. Future iterations should expand to underrepresented regions and rural areas with sparser web documentation to test retrieval under severe information scarcity.
    \item \textit{Addressing Data Contamination:} As with any benchmark derived from the public web, there is a theoretical risk of pre-training data contamination. However, VeriTrip fundamentally mitigates this threat: by forcing agents to resolve fuzzy visual anchors and navigate deliberate corpus noise, and by evaluating them via strict cell-wise factual verification against a frozen sandbox, we heavily penalize pure parametric memorization and ensure success relies on genuine retrieval.
    \item \textit{Interpretive Boundaries:} Finally, VeriTrip evaluates retrieval and multi-constraint planning within a static, evidence-sufficient sandbox. It does not measure the interactive execution capabilities required for real-time, dynamic web booking (e.g., dynamic pricing, or sold-out tickets).
\end{itemize}
%---------------------
{
% \small
\bibliography{neurips}
\bibliographystyle{unsrt}
}
% \section*{References}

% References follow the acknowledgments in the camera-ready paper. Use unnumbered first-level heading for
% the references. Any choice of citation style is acceptable as long as you are
% consistent. It is permissible to reduce the font size to \verb+small+ (9 point)
% when listing the references.
% Note that the Reference section does not count towards the page limit.
% \medskip

% {
% \small

% [1] Alexander, J.A.\ \& Mozer, M.C.\ (1995) Template-based algorithms for
% connectionist rule extraction. In G.\ Tesauro, D.S.\ Touretzky and T.K.\ Leen
% (eds.), {\it Advances in Neural Information Processing Systems 7},
% pp.\ 609--616. Cambridge, MA: MIT Press.

% [2] Bower, J.M.\ \& Beeman, D.\ (1995) {\it The Book of GENESIS: Exploring
%   Realistic Neural Models with the GEneral NEural SImulation System.}  New York:
% TELOS/Springer--Verlag.

% [3] Hasselmo, M.E., Schnell, E.\ \& Barkai, E.\ (1995) Dynamics of learning and
% recall at excitatory recurrent synapses and cholinergic modulation in rat
% hippocampal region CA3. {\it Journal of Neuroscience} {\bf 15}(7):5249-5262.
% }

%%%%%%%%%%%%%%%%%%%%%%%%%%%%%%%%%%%%%%%%%%%%%%%%%%%%%%%%%%%%

\appendix

\clearpage
\setcounter{page}{1}
\setcounter{section}{0}
\setcounter{secnumdepth}{2}
\renewcommand\thesection{\Alph{section}}

This Appendix contains the following sections:
\begin{itemize}
% \item Section~\ref{ap:related_work}: Related Works
% \item  Secion~\ref{ap:limitation} : Limitations
\item Section~\ref{ap:impact}: Societal Impact Statement
\item Section~\ref{ap:benchmark_detail}: Benchmark Details
\begin{itemize}
    \item Subsection~\ref{ap:corpus}: Corpus Construction
    \item Subsection~\ref{ap:VKB}: VKB Construction
    \item Subsection~\ref{ap:query}: Query Construction
\end{itemize}
\item Section~\ref{ap:exp_detail}: Experiment Details
\item Section~\ref{ap:exp_result}: Case Study
\item Section~\ref{ap:prompt}: Prompt List

\end{itemize}

\section{Societal Impact Statement}
\label{ap:impact}
This paper presents work whose goal is to advance the field of machine learning, specifically in the domain of autonomous agents and multimodal information retrieval. 
The broader impacts of our work are summarized as follows:
\begin{itemize}

\item Reliability and Safety in Agent. 
    As Large Language Models are increasingly integrated into real-world decision-making systems, the risk of ``hallucination"—generating plausible but factually incorrect information—poses significant safety and financial risks (e.g., booking incorrect flights or non-existent hotels). 
    VeriTrip directly addresses this challenge by providing a benchmark focused on evidence-based factual grounding. 
    By quantifying the gap between parametric memory and retrieved evidence, our work encourages the development of safer agents that can verify their own actions against trusted sources before execution.
\item The Multimodal Retrieval Base (MRB) is constructed from public web snapshots. 
    We have taken measures to anonymize personally identifiable information (PII) from user-generated content (e.g., social media travel guides) during the data curation process. 
    We emphasize that any deployment of agents based on this research should strictly adhere to privacy regulations and respect the terms of service of data providers.
\item Societal Consequences. 
    By automating complex logistical planning, this technology has the potential to significantly enhance productivity and accessibility for users. 
    However, we acknowledge the potential for economic disruption in the travel consultancy sector. 
    We view our contribution as a step towards human-centric AI collaboration, where agents handle information synthesis to support, rather than replace, human decision-making.
\end{itemize}

\section{Benchmark Details}
\label{ap:benchmark_detail}
\subsection{Corpus Construction}
\label{ap:corpus}
\textbf{Accommodations.} 
For accommodations, we source original data from Ctrip's~\footnote{https://www.ctrip.com/} website.
We extract data spanning from September 1 to September 30, 2025.
We specifically included fields like \textit{Name, Address, Room Types, Price, Reviews, Hotel Facilities, Hotel Policies}. 
Note that hotel dates will not affect the model's search results.

\noindent\textbf{Flights.}
For flights within China, we source original data from Ctrip.
From this website, we extracted data spanning from October 15th to October 30th, 2025.
For flights within America, we extracted data spanning from November 1st to November 14th, 2025.
We specifically include fields like \textit{Flight ID, Flight Date, Departure Time, Arrive Time, Departure Airport, Arrive Airport, Duration, Price.}

\noindent\textbf{Trains.}
The train data is only available in China.
We source original data from 12306 website~\footnote{https://www.12306.cn/index/}.
We extracted data spanning from October 15th to October 30th, 2025.
We specifically included fields like \textit{Train ID, Departure Date, Departure Time, Arrive Time, Departure Station, Arrive Station, Duration, Price of different Seat.}

\noindent\textbf{Attractions.}
We collect information on the top 50-100 attractions in the default recommended order from Ctrip~\footnote{https://www.ctrip.com/} and Trip~\footnote{https://trip.com/}. 
We collected data between September 10th and September 30th.
From this type, we extract \textit{Name, Address, Introduction, Recommended Visit Time, Opening hours, Ticket Price.}

\noindent\textbf{Travel Guides.}
We extract raw data with more than 100 likes from RedNote~\footnote{https://www.xiaohongshu.com/explore}, all of which were posted by real users.
We specifically include fields like \textit{Title, Likes, Content}.
All content consists of text larger than 300 characters.
\subsection{VKB Construction}
\label{ap:VKB}
While the majority of VKB entries (flights, trains, attraction tickets, opening hours, and hotel prices) are rigorously extracted and manually verified directly from the MRB snapshots, we employ auxiliary sources for specific attributes to ensure evaluation precision:

\noindent\textbf{Restaurants.}
Since detailed restaurant information is not readily available website, we use API calls to provide restaurant information, which also serves as data for VKB.
We source original data from Amap API~\cite{amap_api} and Google SerpAPI~\footnote{https://serpapi.com/}.
Essential details such as \textit{Name, Price, Rating, Cuisine, Recommended Food, Area} are provided in this dataset.
Note that the restaurant section in the plan is for preference validation only and does not perform factual checks.

\noindent \textbf{Geospatial Coordinates.}
% In this part, we collect the \textit{latitude, longitude} of all locations(hotels, restaurants and attractions) from Amap API~\cite{amap_api} and Google SerpAPI. 
% Besides, we calculated the Web Mercator projection coordinates using the latitude, longitude.
To support the Geographic Coherence Check and the TSP-based efficiency metric defined in Section ~\ref{sec:eval_protocol}, precise geolocation is required.
We aggregate the \textit{latitude} and \textit{longitude} for all relevant entities (hotels, restaurants, and attractions) via the Amap API~\cite{amap_api} and Google SerpAPI.
To facilitate accurate distance calculations within the solver, we further convert these raw coordinates into Web Mercator projection format.

% \subsection{Query Construction}
% We 

\subsection{Query Construction}
\label{ap:query}
The query generation pipeline follows a ``Seed-and-Expand" paradigm to ensure both realism and diversity.
We first extracted initial queries from existing travel guides within our MRB to serve as ``seed queries."
Subsequently, we employed Qwen3-Max~\cite{yang2025qwen3} to expand and diversify these queries into six distinct queries for each target city, systematically stratified by difficulty level.
% Drawing upon real-world travel guides and the structured data within our MRB, we employed Qwen3-Max~\cite{yang2025qwen3} to synthesize six queries for each target city, stratified by difficulty level. 
% The generation process was constrained by specific parameters to ensure diversity. 
% Travel dates were restricted to the window of October 15th to 30th, 2025, with group sizes ranging from 1 to 8 individuals. Traveler personas were sampled from a set including single travelers, couples, friends, and various family configurations (with elderly, with children, with both, or multi-family groups).
% Furthermore, user preferences were modeled across five dimensions: attractions and dining (mapped to specific entity names within the target city); general interests (selected from cultural history, natural scenery, city charm, leisure and entertainment, or family fun); accommodation (double, twin, single, or family rooms); intercity transportation (high-speed rail or airplane); and inner-city transportation (public transit or taxi). Following the initial generation, we grounded the data by replacing the model-generated attraction and dining preferences with verified entities from our corpus. 
% Finally, the dataset underwent a manual verification and refinement process to ensure logical consistency.
The expansion process was governed by specific constraints:
\begin{itemize}
\item \textbf{Contextual Constraints:} Travel dates were restricted to the MRB snapshot window (October 15th to 30th, 2025). Group sizes ranged from 1 to 8 individuals, covering diverse traveler personas (e.g., solo, couples, multi-generational families).
\item \textbf{Preference Dimensions:} User needs were modeled across five axes: attractions/dining, general interests (e.g., cultural, natural), accommodation type, intercity transport, and inner-city transit.
\item \textbf{Visual Anchors:} To incorporate multimodal challenges, we systematically replaced specific textual mentions of attractions or dining spots with reference images from the corpus, the image has been deliberately cropped to eliminate prominent scenic features while retaining traceable information, simulating real-world ``visual search" intent.
\end{itemize}
Following generation, we grounded the data by ensuring all model-expanded preferences matched verified entities in our corpus.
Finally, the dataset underwent manual verification to ensure that every query is logically consistent and demonstrably solvable using only the frozen MRB.

\section{Experiment Details}
\label{ap:exp_detail}
\subsection{Baselines}
    The specific version of MLLMs:
    GPT-5-mini(gpt-5-mini-2025-08-07)~\cite{gpt5}, GPT-4o (gpt-4o-2024-08-06)~\cite{hurst2024gpt}, GPT-4o-mini (gpt-4o-mini-2024-07-18)~\cite{gpt4omini}, Gemini-2.5-flash (gemini-2.5-flash-2025-06-17)~\cite{comanici2025gemini},  Claude-4.5-Sonnet (claude-sonnet-4-5-20250929)~\cite{claude4_5}, Claude-3.7-Sonnet (claude-3-7-sonnet-20250219)~\cite{claude3_7}, Qwen-VL-Max~\cite{qwen3-vl-max}, Qwen3-VL-235b~\cite{yang2025qwen3}, o3 (o3-2025-04-16)~\cite{gpto3}, o4-mini (o4-mini-2025-04-16)~\cite{gpto4mini}, Gemini-2.5-pro (gemini-2.5-pro-2025-06-17)~\cite{comanici2025gemini}.
    
\subsection{Implementation Details}
We set MAX$\_$TOKENS=16384 as the default for models without thinking, and 32768 for thinking models.
The thinking mode is moderate.
MLLM autonomously chooses whether to invoke tools and when to invoke which tool.
For the Qwen3-Embedding-8b retriever, we set the passage max length to 4096 and the per-device eval batch size to 32.
For the search tool, we used Faiss~\footnote{https://github.com/texttron/tevatron}, with snippet-max-tokens set to 100, and returned the first k=10 snippets.

\subsection{Constraint List}
\label{ap:constraints}
Following TravelPlanner~\cite{xie2024travelplanner} and ChinaTravel~\cite{shao2024chinatravel}, we utilize commonsense constraints to evaluate whether agent can incorporate commonsense into their plan without explicit instructions.
Unlike previous methods, which categorize constraints into commonsense and hard constraints, we believe that a qualified travel plan should simultaneously satisfy the constraints listed in Table~\ref{ap_tab:constraint}.

\begin{table}[ht]
\centering
\small
\caption{Constraint and preference description. The planning constraints are evaluated based on how well the MLLM agents' plan aligns with these specific rules.}
\begin{tabular}{l p{9cm}}
\toprule
Constraint                   & Description \\
\midrule
is\_within\_budget & Within the budget evaluation, the calculated total cost of the plan must be within the user's specified budget (allowing for a 10$\%$ tolerance)\\ \midrule
first\_activity\_after\_arrival &Within the itinerary timing evaluation, the first activity (that is not inter-city transport) must start at or after the arrival time of the first transportation leg. \\ \midrule
last\_activity\_before\_departure & Within the itinerary timing evaluation, the last activity (that is not inter-city transport) must end at or before the departure time of the final transportation leg. \\ \midrule
transportation\_continuity& Within the transportation evaluation, the arrival station of each inter-city leg must match the departure station of the subsequent leg. \\ \midrule
transportation\_closed\_loop&  Within the transportation evaluation, the final arrival city of the last transportation leg must match the initial departure city of the first leg. \\ \midrule
transportation\_dates& Within the transportation evaluation, the departure date of the first leg and the return date of the last leg must exactly match the user's query start and end dates \\ \midrule
accommodation\_capacity& Within the accommodation evaluation, the total capacity of all planned rooms must be sufficient for the number of people specified in the query. \\ \midrule
accommodation\_coverage& Within the accommodation evaluation, the accommodation plan must continuously cover the entire travel period, with check-in/check-out dates aligning with the trip's start/end dates and having no gaps between bookings. \\ \midrule
attractions\_uniqueness& Within the attraction evaluation, all planned attractions must be unique (not visited more than once). \\ \midrule
meals\_uniqueness& Within the meal evaluation, all planned restaurants (excluding breakfast) must be unique. \\ \midrule
inner\_city\_time\_efficiency& Within the inner-city transportation evaluation, the ratio of commuting time to total daily activity time must not exceed a set limit (35$\%$) on full travel days. \\ \midrule

days\_coverage\_correct& Within the day evaluation, the set of dates with planned activities must exactly match the date range specified in the user's query.\\ \midrule
activity\_presence\_on\_mid\_days& Within the day evaluation, all intermediate travel days (not the first or last day) must include at least one `meal' and one `attraction'.\\ \midrule
information\_density & Within the day evaluation, the total number of planned activities must meet a minimum information density threshold relative to the trip duration. \\ \midrule
\rowcolor{gray!25}\multicolumn{2}{c}{Preference} \\
\midrule
intercity\_transportation\_preference & Within the transportation evaluation, the planned mode of inter-city transport (e.g., plane, train) must match the user's stated preference. \\ \midrule
accommodation\_preference& Within the accommodation evaluation, at least one of the planned room types should match the user's preference if one was specified. 
\\ \midrule
attraction\_preference& Within the attraction evaluation, all `must-see' attractions specified by the user must be included in the plan. \\ \midrule
meal\_preference& Within the meal evaluation, the plan must include restaurants or meal types that match the user's stated preferences. \\ \midrule
inner\_city\_transportation\_preference& Within the inner-city transportation evaluation, the planned modes of transport must not conflict with the user's stated preference (e.g., planning a taxi when public transport was preferred). \\ 
\bottomrule
\end{tabular}
\label{ap_tab:constraint}
\end{table}

\begin{figure}[ht]
    \centering
    \begin{subfigure}{\columnwidth}
        \centering 
        \includegraphics[width=0.6\linewidth]{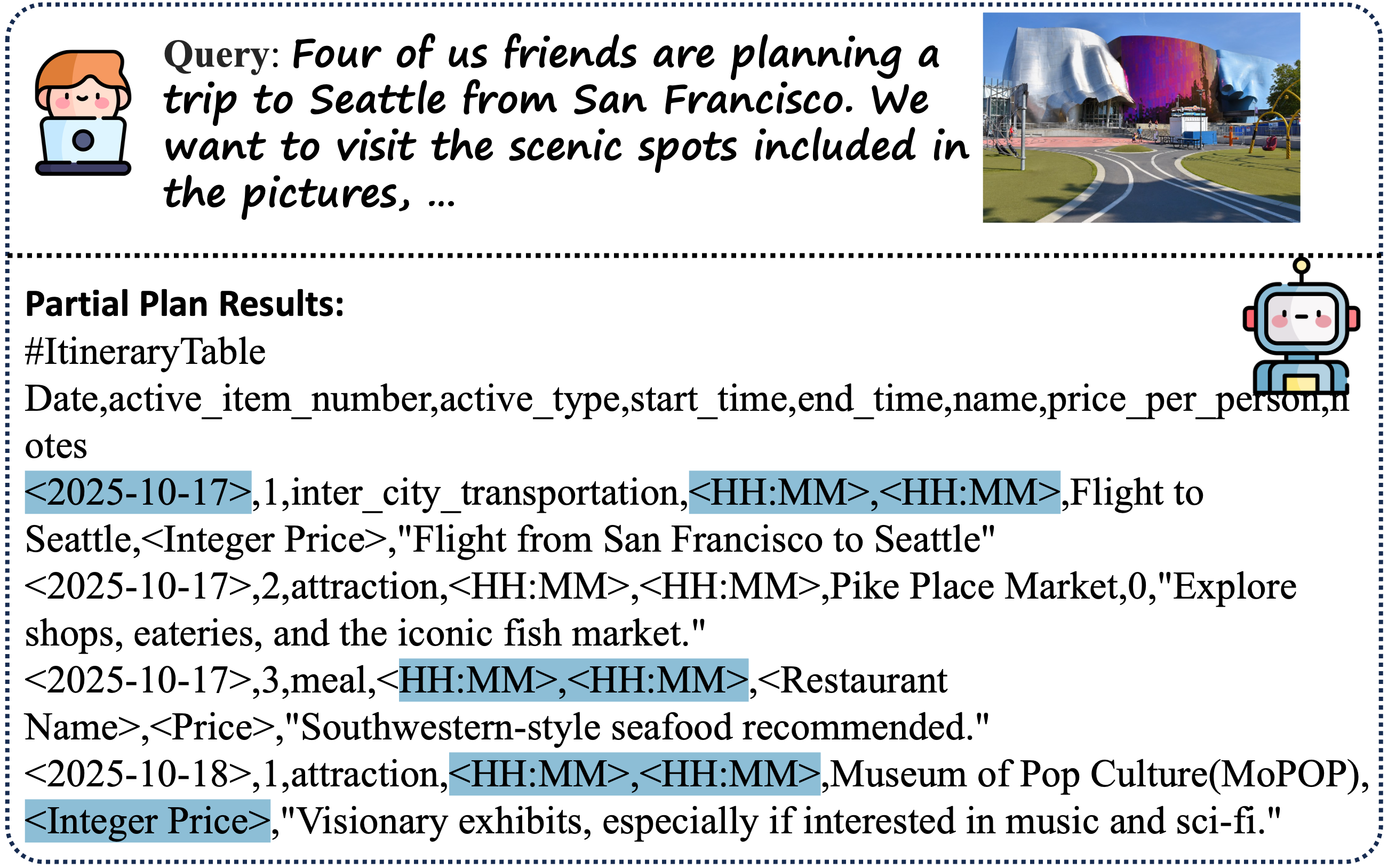}
        \caption{Format error case of GPT-4o.
        }
        \label{fig:sub1}
    \end{subfigure}
    \vspace{4mm}
    \begin{subfigure}{\columnwidth} 
        \centering
        \includegraphics[width=0.6\linewidth]{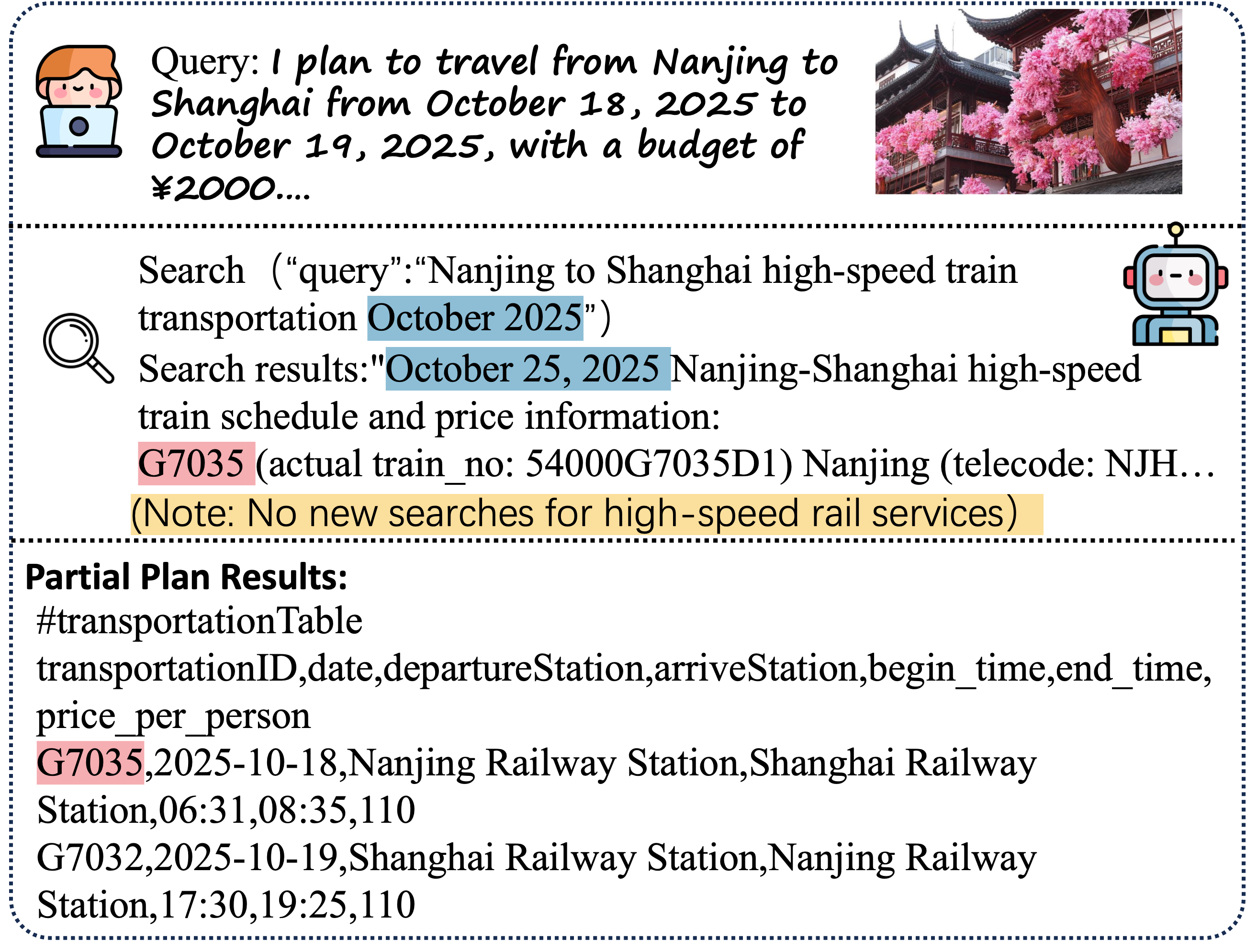}
        \caption{Fact error case of Qwen-VL-max. 
        }
        \label{fig:sub2}
    \end{subfigure}
    \caption{Case Study of failures. (a) The plan failed because it copied the formatting requirements from the predefined prompt. (b) After performing a search, the agent obtained relevant high-speed rail information that did not match the specific date. Instead of re-initiating a search for the target date, the agent directly added the high-speed rail number to the plan, causing all FR indicators for this high-speed rail information to be judged as incorrect.}
    \label{fig:case} 
\end{figure}

\begin{figure}[ht]
    \centering
    \begin{subfigure}{\columnwidth}
        \centering 
        \includegraphics[width=0.6\linewidth]{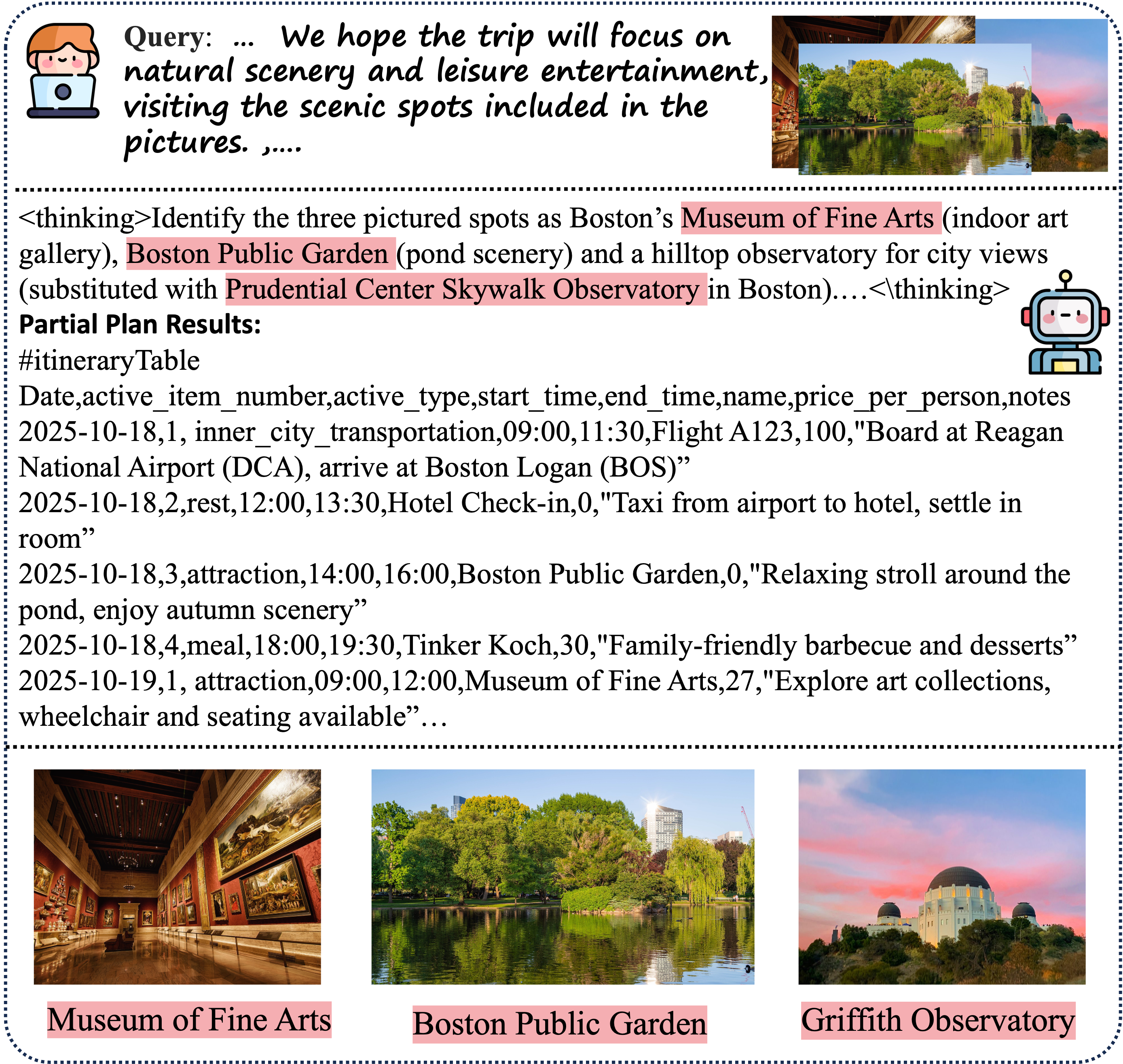}
        \caption{Image comprehension error case 1 from o3.
        }
        \label{ap:subfig1}
    \end{subfigure}
    \vspace{4mm}
    \begin{subfigure}{\columnwidth} 
        \centering
        \includegraphics[width=0.6\linewidth]{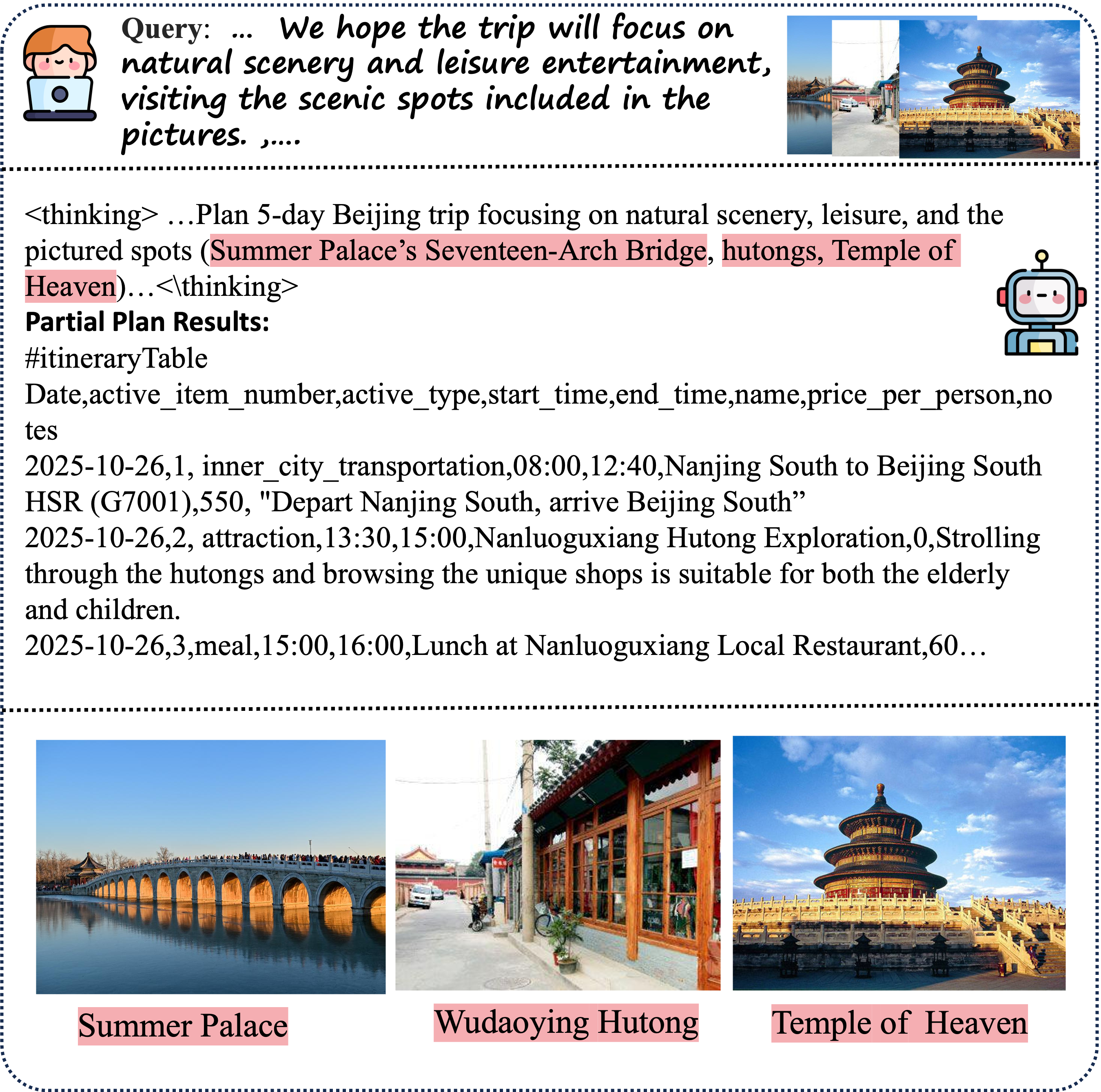}
        \caption{Image comprehension error case 2 from o3.
        }
        \label{ap:subfig2}
    \end{subfigure}
    \vspace{-8mm}
    \caption{Case Study of failures.  These plans failed because of errors or insufficient identification of certain landmarks during the image understanding phase.}
    \label{ap:fig_case} 
\end{figure}

\begin{figure}[ht]
\centering 
\includegraphics[width=0.6\linewidth]{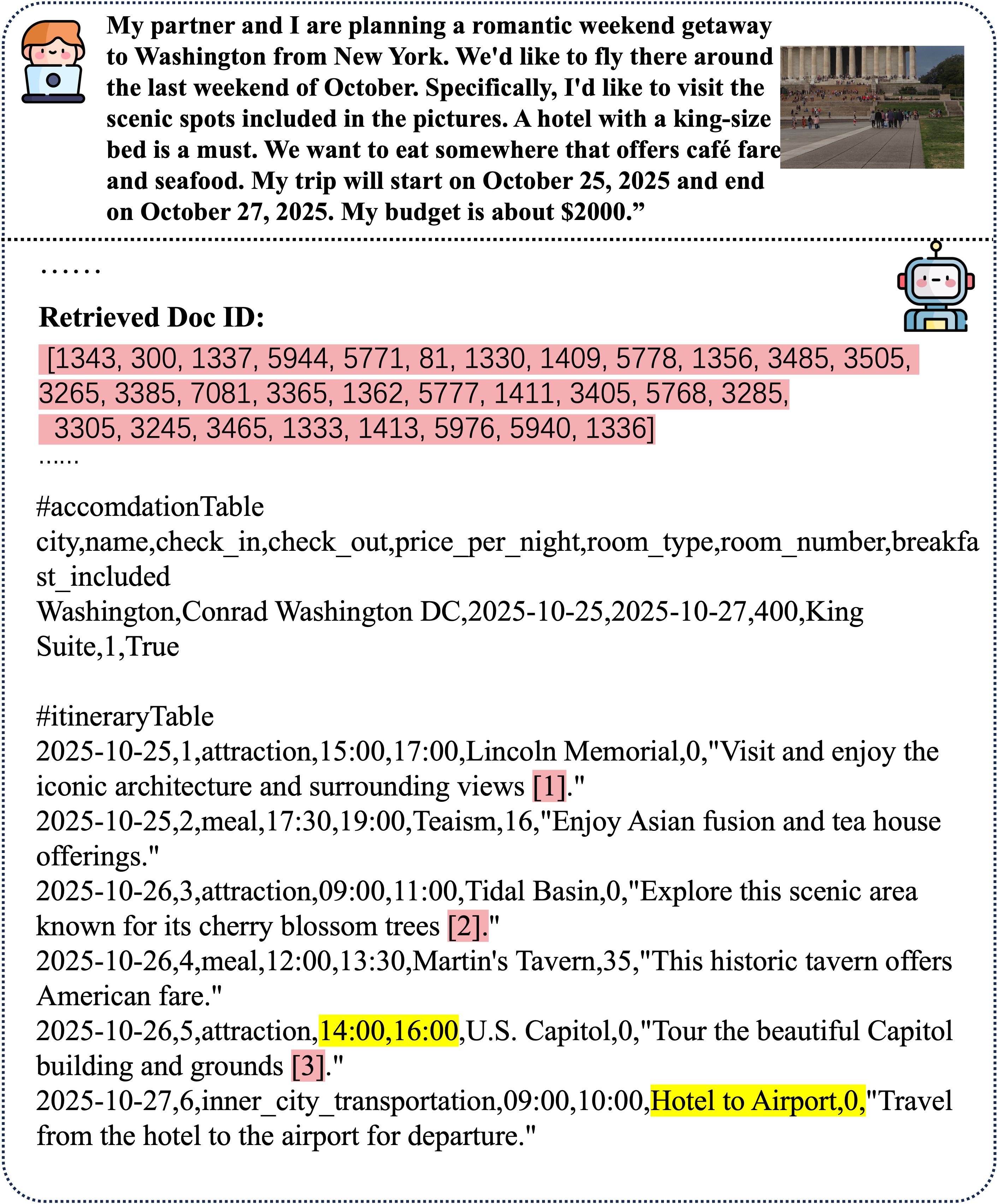}
\caption{
This is an example of hallucination in the evidence document generated by GPT-4o-mini. Although the tool returned the retrieved document IDs to the model, the model lazily used simple numbers 1, 2, and 3 for labeling. We have also highlighted in yellow the illogical outputs, such as the day's itinerary ending at 16:00 and the cost of traveling from the hotel to the airport being 0.
}
\label{fig:docid}
\end{figure}

\section{Case Study}
\label{ap:exp_result}

We present some classic failure cases as shown in Figure~\ref{fig:case}.
A critical failure mode involves the agent's inability to distinguish between task instructions and content generation.
This indicates a fundamental weakness in interpreting meta-instructions.
Another prevalent issue is a lack of rigorous verification during the retrieval process.
Instead of initiating a corrective search to find data matching all constraints, the agent prematurely accepts and incorporates the incorrect detail into the final plan. 
This failure demonstrates a deficiency in iterative refinement, where the agent settles for plausible-but-unverified information rather than ensuring every detail is factually correct, resulting in a cascade of grounding errors.

\subsection{Failure Case Studies}
Despite the promising performance of MLLMs in travel planning tasks, our qualitative analysis reveals critical robustness issues in visual scene understanding and entity grounding. Figure~\ref{ap:fig_case} illustrates two representative failure cases where the MLLM-based agent fails to accurately interpret user intent derived from visual inputs.

\noindent 1. Contextual Bias and Visual Hallucination

The most significant failure mode observed is contextual hallucination, where the model’s strong textual priors override visual evidence.
In the Boston itinerary query(Figure~\ref{ap:subfig1}), the user provided three images: the Museum of Fine Arts, the Boston Public Garden, and the Griffith Observatory (a landmark located in Los Angeles, not Boston). 
The user's intent was likely to find a similar type of attraction in Boston or was an adversarial input containing a geographical conflict.
The MLLM failed to detect the geographical inconsistency. Instead of identifying the third image as the Griffith Observatory and flagging the conflict, the model recognized the semantic category of the image (an observatory/viewpoint) and forced it to fit the target city constraint (Boston). 
Consequently, it hallucinated a correspondence to the Prudential Center Skywalk Observatory.
This error suggests that current MLLMs suffer from contextual over-reliance. 
The explicit textual instruction (planning a trip for ``Boston") acts as a strong prior that suppresses the visual recognition of out-of-distribution entities. 
The model prioritizes logical consistency in the text generation (listing a Boston observatory) over faithful visual grounding.

\noindent  2. Limitations in Fine-Grained Entity Recognition

The second failure mode highlights the trade-off between generic scene recognition and specific entity linking(Figure~\ref{ap:subfig2}).
The user provided an image of Wudaoying Hutong—a specific, culturally significant alley in Beijing known for its distinct architecture and shops.
While the model correctly identified the general scene category as ``hutongs", it failed to recognize the specific instance (``Wudaoying"). The resulting recommendation was a generic visit to ``hutongs".
This indicates a limitation in fine-grained visual recognition. While MLLMs are proficient at zero-shot classification of general objects (e.g., ``bridge", ``alley", ``temple"), they struggle with specific landmark identification when the visual features are subtle or require domain-specific knowledge (knowledge cutoff or training data sparsity). For a travel agent, this lack of granularity leads to suboptimal planning, as generic advice fails to capture the specific atmosphere or logistics implied by the user's reference image.

\noindent  3. The document ID was generated by the hallucination.

As shown in Figure~\ref{fig:docid}, although the tool has returned the retrieved IDs for the model, the model lazily uses simple numbers 1, 2, and 3 for labeling instead.

\section{Prompt List}
\label{ap:prompt}
\subsection{SYSTEM PROMPT}
We provide the system prompt of agents as follows:
\begin{tcolorbox}[title= SYSTEM\_PROMPT,colback=gray!15,colframe=gray!50!black,arc=1mm,boxrule=1pt,left=1mm,right=1mm,top=1mm,bottom=1mm, breakable]
\small
$\#$ Role Setting \\
You are a professional and meticulous travel information collection and organization expert. 
You are capable of fully understanding user needs, skillfully using search tools, and planning travel itineraries that thoroughly consider user requirements. \\
$\#$ Task Description \\
Upon receiving a user's question, you must use the provided tools to separately search for information on the destination's travel guides, high-speed trains or flights, hotels, attractions, and restaurants. 
You will then use the information obtained from these tools to create a travel plan.
To help you complete your tasks efficiently and accurately, you have been provided with 6 tools. 
You **must** call these tools to generate the travel plan and **must not** answer based on your existing knowledge.
You can break down the task into multiple sub-tasks as needed and use the tools multiple times to ensure you find the necessary answers within the information provided by the tools.
If you believe the information obtained from the tool call is incomplete or incorrect, making it insufficient to answer the user's question, you must think about what other information is needed and call the tool again. 
Continue this process until the collected information is sufficient to complete the user's requested travel plan.

\end{tcolorbox}

\subsection{SEARCH PROMPT}
We provide the search prompt for the React strategy as follows:

\begin{tcolorbox}[title= SEARCH\_PROMPT,colback=gray!15,colframe=gray!50!black,arc=1mm,boxrule=1pt,left=1mm,right=1mm,top=1mm,bottom=1mm, breakable]
\small
Question: \{Question\} \\
Please collect information on travel guides, transportation, dining, attractions, and accommodation based on user needs, and plan a travel itinerary as detailed as possible.
The output must be valid JSON containing exactly three top-level keys: transportationTable, accommodationTable, and itineraryTable. 
Do not include any extra fields, explanatory text, Markdown syntax, or comments.
\begin{lstlisting}[language=Python][frame=lines, framesep=2mm, breaklines=true, fontsize=\small]{json}

{
    "Final Result": {
        "transportationTable": [
            {
                "transportationID": "<Flight ID/Train ID>",
                "date": "<YYYY-MM-DD>",
                "departureStation": "<Departure City Airport/High-Speed Rail Station>",
                "arriveStation": "<Arrival City Airport/High-Speed Rail Station>",
                "begin_time": "HH:MM ",
                "end_time": "HH:MM ",
                "price_per_person": "<Integer Price>"
            },
            {
                "transportationID": "<Flight ID/Train ID>",
                "date": "<YYYY-MM-DD>",
                "departureStation": "<Arrival City Airport/High-Speed Rail Station>",
                "arriveStation": "<Departure City Airport/High-Speed Rail Station>",
                "begin_time": "HH:MM ",
                "end_time": "HH:MM ",
                "price_per_person": "<Integer Price>"
            }
        ],
        "accommodationTable": [
            {
                "city": "<City Name>",
                "name": "<Hotel Name>",
                "check_in": "<YYYY-MM-DD>",
                "check_out": "<YYYY-MM-DD>",
                "price_per_night": "<Integer Price>",
                "room_type": "<Room Type>",
                "room_number": "<Number of Rooms>",
                "breakfast_included": "<true/false>"
            }
        ],
        "itineraryTable": [
            {
                "Date": "<YYYY-MM-DD>",
                "active_item_number": "<Serial Number>",
                "active_type": "<Activity Type>",
                "start_time": "HH:MM ",
                "end_time": "HH:MM ",
                "name": "<Activity Name>",
                "price_per_person": "<Integer Price>",
                "notes": "<Detailed Activity Notes>"
            },
            {
                "Date": "<YYYY-MM-DD>",
                "active_item_number": "<Serial Number>",
                "active_type": "<Activity Type>",
                "start_time": "HH:MM ",
                "end_time": "HH:MM ",
                "name": "<Activity Name>",
                "price_per_person": "<Integer Price>",
                "notes": "<Detailed Activity Notes>"
            }
        ]
    }
}
\end{lstlisting}
itineraryTable Rules:
The value for the \texttt{active\_type} field must be one of the following four: `inner\_city\_transportation', `inter\_city\_transportation', `meal', or `attraction'.
For attraction type, you should cite your evidence documents by enclosing in square brackets in its notes, for example:[20].
For \texttt{inner\_city\_transportation}, the name should simply state the origin and destination (e.g., `Hotel to Museum'). 
For activities of type meal, the name should be the restaurant's name. For hotel breakfast, it should be the hotel's name. The notes field should provide recommendations for the restaurant's cuisine.

Field Format:
All dates must be in \texttt{YYYY-MM-DD} format, and all times must be in \texttt{HH:MM} format.
All prices (\texttt{price\_per\_person}, \texttt{price\_per\_night}) must be integers.
\end{tcolorbox}

\subsection{QUERY GENERATION PROMPT}
The instruction prompt for query generation is provided as follows:
\begin{tcolorbox}[title= QUERY\_GENERATION\_PROMPT, colback=gray!15,colframe=gray!50!black,arc=1mm,boxrule=1pt,left=1mm,right=1mm,top=1mm,bottom=1mm, breakable]
\begin{lstlisting}[language=Python][frame=lines, framesep=2mm, breaklines=true, fontsize=\small]{json}
    "query_example" : {
    "id": "query1", 
    "start_city": "Guangzhou", 
    "target_city": "Hangzhou", 
    "start_date": "2025-10-28", 
    "end_date": "2025-10-30", 
    "people_number": 1, 
    "budget": 3000, 
    "persona": "Solo traveler",
    "have_pet": false, 
    "difficulty": "Medium", 
    "p_attraction": "West Lake Scenic Area|Lingyin Temple", 
    "p_meal": "Longjing Shrimp|West Lake Vinegar Fish", 
    "p_accommodation": "Double room", 
    "p_interest": "Cultural history", 
    "p_intercity_traffic": "High-speed rail", 
    "p_innercity_traffic": null, 
    "nature_language": "I am a photography enthusiast planning a solo high-speed rail trip from Guangzhou to Hangzhou from 2025-10-28 to 2025-10-30. I enjoy cultural history. My budget is 3000. I hope to visit the West Lake Scenic Area and Lingyin Temple, and taste Longjing Shrimp and West Lake Vinegar Fish. I prefer a hotel with a double bed. Please help me plan a relaxed, go-with-the-flow itinerary."
}
\end{lstlisting}
\small
You are a travel planning expert. Based on the reference information, randomly design 6 travel queries categorized as Easy, Medium, and Hard. The query format must reference {query\_example}, and the keys in the output must include the fields appearing in that format.

Specific constraints are as follows:
- start\_city is {i}, target\_city is {j};
- start\_date and end\_date must be within the range of 2025-10-15 to 2025-10-30;
- people\_number is within [1, 2, 3, 4, 5, 6, 7, 8];
- persona is selected from [Solo traveler, Couple, Friends, Family (with elderly), Family (with children), Family (with elderly and children), Two families];
- p\_attraction refers to attraction names within the target\_city;
- p\_meal refers to specialty foods or restaurant names within the target\_city;
- p\_interest is selected from [Cultural history, Natural scenery, City charm, Leisure and entertainment, Family fun];
- p\_accommodation is selected from [Double room, Twin room, Single room, Family room];
- p\_intercity\_traffic is selected from [High-speed rail, Airplane];
- p\_innercity\_traffic is selected from [Public transportation, Taxi].

Constraints for generation:
- The 'nature\_language' field does not need to contain the value information for every key listed above; however, field information missing from the natural language must still exist in the JSON structure to ensure format consistency across all queries. You may use None to represent missing values.
- p\_attraction, p\_meal, p\_accommodation, p\_interest, and p\_innercity\_traffic represent user preferences. If a value is empty, write None. The value for each preference category must not exceed 3 items.
- Add user preference fields according to the different difficulty levels. 
    - Easy queries: 2-3 preferences.
    - Medium queries: 3-4 preferences.
    - Hard queries: More than 4 preferences.
- The difficulty of the query should also increase with the complexity of 'people\_number' and 'persona'.
- The travel requirements in the query must align with common sense and the specific needs of different personas (e.g., relaxed/laid-back, high-intensity/packed itinerary, fulfilled but elderly-friendly, etc.).

Please output the questions you designed in JSON format.

\end{tcolorbox}

%%%%%%%%%%%%%%%%%%%%%%%%%%%%%%%%%%%%%%%%%%%%%%%%%%%%%%%%%%%%

% \newpage
% \input{checklist.tex}

\end{document}